\PassOptionsToPackage{table}{xcolor}
\documentclass{article} % For LaTeX2e
\usepackage{iclr2027_conference,times}

\usepackage[utf8]{inputenc}
\usepackage[T1]{fontenc}
\usepackage{booktabs}
\usepackage{nicefrac}
\usepackage{microtype}
\usepackage{amssymb}
\usepackage{xcolor}
\usepackage{graphicx}
\usepackage{algorithm}
\usepackage{algorithmic}
\usepackage{float}
\usepackage{wrapfig}

\newcommand{\gain}[1]{\textcolor{green!55!black}{$\uparrow$\,#1}}
\newcommand{\drop}[1]{\textcolor{red}{$\downarrow$\,#1}}
\makeatletter
\newcommand{\minitablecaption}[1]{\def\@captype{table}\caption{#1}}
\makeatother

\usepackage{amsmath,amsfonts,bm}

\def\eqref#1{equation~\ref{#1}}
\def\1{\bm{1}}

\DeclareMathAlphabet{\mathsfit}{\encodingdefault}{\sfdefault}{m}{sl}
\SetMathAlphabet{\mathsfit}{bold}{\encodingdefault}{\sfdefault}{bx}{n}

\usepackage{hyperref}
\usepackage{url}

\IfFileExists{fontawesome5.sty}{%
  \usepackage{fontawesome5}%
  \newcommand{\githubicon}{\faGithub}%
}{%
  \newcommand{\githubicon}{}%
}
\newcommand{\checkdata}[3][\githubicon]{\unskip\thanks{#1\ #2:~\href{#3}{\texttt{#3}}}}

\title{\texorpdfstring{P$^2$O}{P2O}: Joint Policy and Prompt Optimization}

\author{%
  Xinyu Lu$^{\ 1,2}$\thanks{\textbf{Equal contribution}}
  \checkdata[\githubicon]{Code}{https://github.com/zlkqz/P2O}
  \And Kaiqi Zhang$^{1,2 *}$ \And Jinglin Yang$^{3,4,5}$ \And Boxi Cao$^{1}$  \And Yaojie Lu$^{1}$ \AND Hongyu Lin$^{1}$ \hfill Min He$^{5}$ \hfill Xianpei Han$^{1,2}$ \hfill Le Sun$^{1,2}$ \vspace{0.5cm}
  \\
  $^{1}$Chinese Information Processing Laboratory, Institute of Software, Chinese Academy of Sciences \\
  $^{2}$University of Chinese Academy of Sciences \\
  $^{3}$Institute of Information Engineering, Chinese Academy of Sciences \\
  $^{4}$School of Cyber Security, University of Chinese Academy of Sciences\\
  $^{5}$National Computer Network Emergency Response Technical Team/Coordination Center of China \\
  \texttt{zhangkaiqi.zlk@gmail.com, luxinyu2021@iscas.ac.cn}
}

\iclrfinalcopy % Uncomment for camera-ready version, but NOT for submission.
\begin{document}

\maketitle

% ===== ARXIV-ONLY-BEGIN =====================================================
% The style file prints the ICLR notice into the running header inside
% \maketitle, so the logo has to be installed after \maketitle.
\fancyhead[L]{\raisebox{2pt}{\includegraphics[height=26pt]{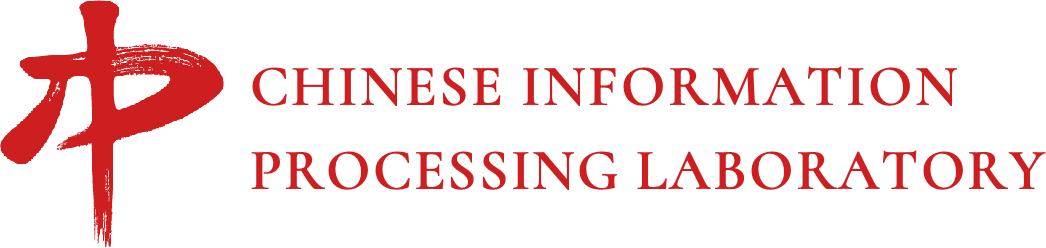}}}
% ===== ARXIV-ONLY-END =======================================================

\begin{abstract}

Reinforcement Learning with Verifiable Rewards (RLVR) enhances Large Language Model (LLM) reasoning but is suffer from advantage collapse: when all rollouts of a query receive identical rewards, the group variance vanishes, most damagingly on hard samples, where scaling rollout budgets yields little. We introduce Joint Policy and Prompt Optimization (P$^2$O) to mitigate this collapse by alternating continuous policy updates with discrete prompt evolution. P$^2$O mines hard samples with a success-rate threshold, evolves reasoning prompts for them with GEPA, and internalizes the elicited trajectories via context distillation, which optimizes each trajectory under the original query and thus removes inference-time prompting, with a Context Ratio Mask (CRM) filtering out extreme likelihood ratios. P$^2$O restores critical advantage signals and surpasses the GRPO baseline by up to 8.2 points in average accuracy on six held-out benchmarks across all training datasets and backbones, while also outperforming DAPO and other baselines. The gains are especially pronounced on hard benchmarks, reaching up to 16.3 points above GRPO on average across AIME24 and AIME25. Our findings expose the limits of standard exploration in sparse-reward environments, illuminating the potential of unifying evolutionary algorithms with reinforcement learning. This integration of discrete semantic search and continuous parameter updates provides a self-reinforcing framework that facilitates more effective LLM alignment.

\end{abstract}

\section{Introduction}

Large Language Models (LLMs) have achieved remarkable proficiency across a spectrum of complex reasoning tasks~\citep{jaech2024openai,guo2025deepseek}.
A defining characteristic of these tasks is the existence of objective verification mechanisms capable of providing deterministic feedback.
Capitalizing on this, Reinforcement Learning with Verifiable Rewards (RLVR)~\citep{lambert2024tulu} has emerged as a dominant paradigm for reasoning alignment.
By leveraging outcome-based supervision, RLVR enables models to autonomously explore the solution space, thereby bypassing the inherent bottlenecks of imitation learning~\citep{liu2025prorl}.

Despite its potential, RLVR is fundamentally bottlenecked by \textit{advantage collapse}: when all rollouts within a group exhibit zero reward variance---whether uniformly succeeding on trivial instances or uniformly failing on challenging ones---the advantage estimates vanish, thereby neutralizing the learning signal~\citep{song2025outcome}. This failure mode on hard samples is particularly detrimental; although these instances carry high informational value, they yield no effective supervision. Moreover, scaling rollout budgets offers fundamentally limited gains, extensive sampling cannot rescue a universally failing policy from zero-reward trajectories.

Common approaches to mitigate exploration challenges include curriculum learning strategies, which progressively introduce harder samples as training advances~\citep{blakeman2025nemotron}, and various reward shaping techniques that provide intermediate feedback.
However, curriculum learning requires heuristic-based and computationally expensive schedule generation~\citep{bercovich2025llama}, while reward shaping demands domain-specific expert heuristics.

% \newpage
\begin{wrapfigure}[24]{r}{0.50\textwidth}
  \vspace{-0.8\baselineskip}
  \centering
  \includegraphics[width=0.48\textwidth]{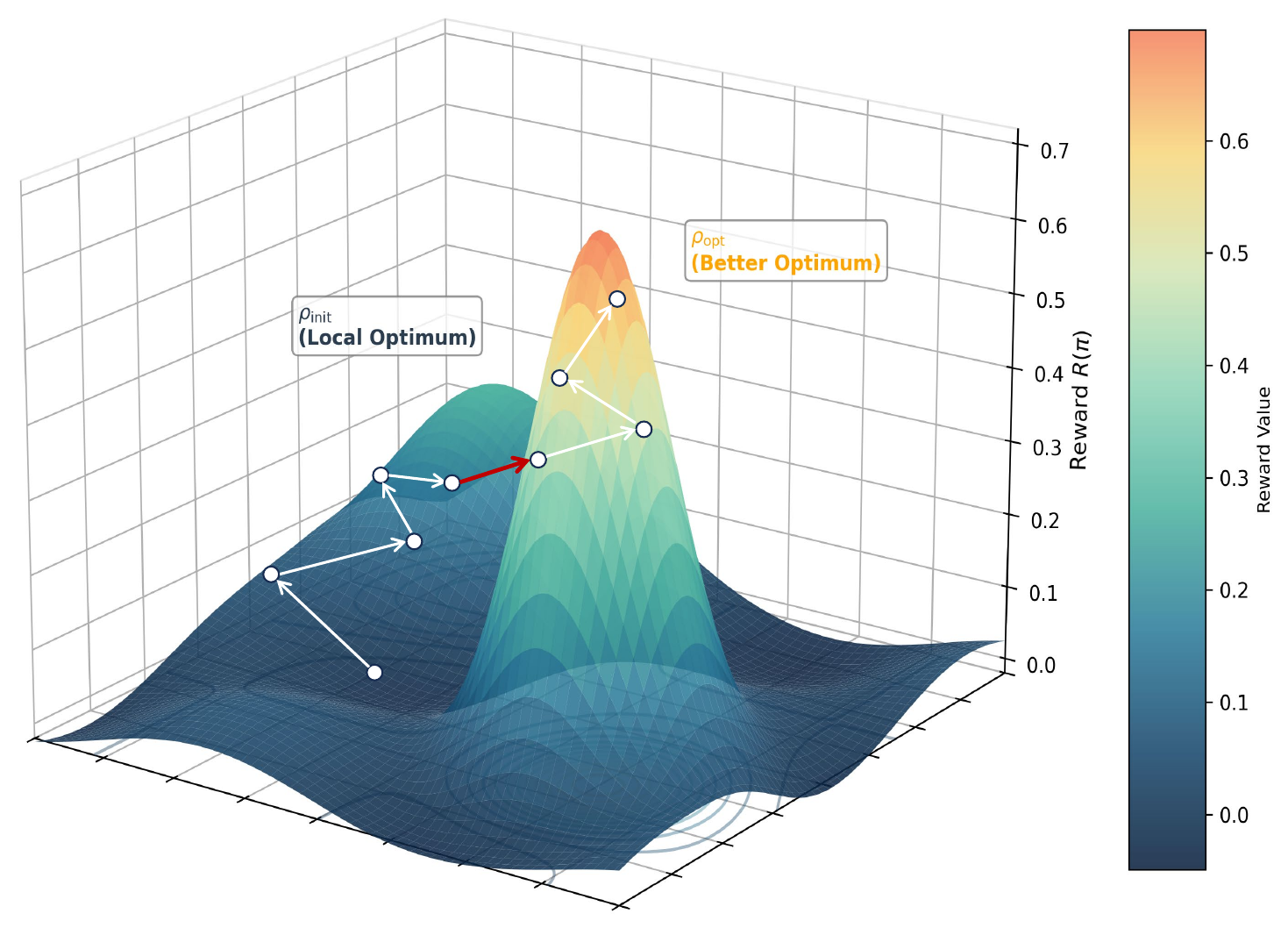}
  \caption{\textbf{Conceptual Illustration of the P$^2$O Framework.} Standard policy optimization often gets trapped in local optima ($\rho_{\text{init}}$) due to sparse rewards on hard samples. P$^2$O bridges this exploration gap using optimized prompts (the red arrow) to reach high-reward regions that are inaccessible via standard exploration. Subsequently, the model consolidates these gains (the white arrows) by updating its parameters to master the new region ($\rho_{\text{opt}}$), effectively internalizing the prompt-induced capabilities.}
  \label{fig:concept}
  \vspace{-0.5\baselineskip}
\end{wrapfigure}

The recent success of prompt optimization methods~\citep{fernando2023promptbreeder,opsahl-ong-etal-2024-optimizing,yuksekgonul2025optimizing} offers a compelling way to break this stalemate.
These methods demonstrate that even when a model fails to solve a hard problem under a standard policy, a carefully evolved prompt can often elicit the correct reasoning path.
This implies that the solution lies within the model's latent search space but is inaccessible via standard gradient ascent~\citep{zelikman2022star}.
As illustrated in Figure~\ref{fig:concept}, optimized prompts act as a bridge, enabling the model to ``jump'' out of the local optimum ($\rho_{\text{init}}$) and cross the reward-sparse valley.
However, relying solely on inference-time prompts is insufficient. The ``jump'' must be followed by internalization, where the model parameters are updated to master the new region ($\rho_{\text{opt}}$).

Building on this intuition, we propose P$^2$O (Joint \textbf{P}olicy and \textbf{P}rompt \textbf{O}ptimization), a novel framework that synergizes adaptive prompt evolution with reinforcement learning to overcome the hard sample bottleneck.
Specifically, P$^2$O identifies challenging instances through an empirical success-rate threshold and employs the Genetic-Pareto (GEPA)~\citep{agrawal2025gepa} prompt optimization algorithm to evolve templates that elicit successful reasoning chains for them.
Rather than relying on inference-time prompting, we distill these trajectories into the policy by scoring them under the original query with a stop-gradient behavior source, while our Context Ratio Mask filters out tokens whose likelihood departs too far from the original-query distribution.
This creates a self-reinforcing cycle: as the model improves, previously hard samples become tractable, and GEPA focuses its efforts on the new frontier of difficulty.

The core contributions of this paper include:

(1) We propose P$^2$O, which alternates continuous policy optimization with discrete prompt evolution, restoring non-degenerate advantage estimates on the hard samples where standard RLVR receives insufficient learning signals.

(2) We introduce a context distillation strategy with a Context Ratio Mask (CRM) that distills reasoning gains induced by optimized prompts directly into model parameters, eliminating inference-time prompt dependency and stabilizing optimization at scale.

(3) We empirically demonstrate that P$^2$O breaks the rollout-scaling ceiling of vanilla GRPO: it surpasses GRPO, DAPO, and compute-matched reflection and distillation baselines without any external teacher, and its effectiveness stems from successful parameter internalization rather than the sheer capability of the reflection operator.

\section{Preliminaries}

\subsection{Policy Optimization}

\label{sec:policy_optimization}

In RLVR, the LLM is parameterized as a policy $\pi_\theta$ that generates a response $y$ for a query $x \in \mathcal{D}$, and a verifier provides the reward $r(x,y)$. We use GRPO~\citep{shao2024deepseekmath}, which avoids a separate critic by estimating the baseline from a group of $K$ responses sampled for the same query. For a query $x$, GRPO samples a response group $\{y_i\}_{i=1}^{K}$ and computes the relative advantage of each response as:

\begin{equation}
A(x,y_i)=
\frac{r(x,y_i)-\mu_{\mathrm{group}}}
{\sigma_{\mathrm{group}}},
\end{equation}

where $\mu_{\mathrm{group}}$ and $\sigma_{\mathrm{group}}$ denote the mean and standard deviation of rewards within the sampled group. When all responses in a group receive the same binary reward, the group variance becomes zero, causing the normalized advantage estimate to degenerate and providing no effective learning signal. This advantage collapse motivates our intervention on hard samples.

\subsection{Prompt Optimization}
\paragraph{Optimization Objective.}
Prompt optimization treats the LLM as a frozen black-box function $\mathcal{M}$ and searches over discrete prompts $z$ in the space of natural-language strings $\mathcal{P}$. Given an input $x$, the model produces $\hat y=\mathcal{M}(z,x)$, and the optimization problem is formulated as:
\begin{equation}
z^* = \arg\max_{z \in \mathcal{P}} \mathbb{E}_{(x, y) \sim \mathcal{D}_{\text{val}}} \left[ S(\mathcal{M}(z, x), y) \right]
\end{equation}
Here $S(\cdot,\cdot)$ denotes the task metric.
Unlike policy optimization, which adjusts continuous weights $\theta$, this is a discrete, derivative-free search over the semantic space of instructions.
\paragraph{GEPA (Genetic-Pareto).}
We use GEPA~\citep{agrawal2025gepa} as the prompt optimizer. GEPA uses a reflection model to turn error feedback into semantic mutations and maintains a diverse candidate population through Pareto-based selection. Its specific adaptation is detailed in the prompt-evolution procedure in Section~\ref{sec:method}.

\section{P$^2$O: Joint Policy and Prompt Optimization}
\label{sec:method}

Policy optimization and prompt optimization act on complementary spaces. P$^2$O couples them by introducing a set of prompt templates $\mathcal{Z}$ for samples on which ordinary policy rollouts provide little learning signal. Let $\mathcal{D}_{\text{hard}} \subset \mathcal{D}$ denote this subset and let $\mathcal{T}(x,z)$ insert template $z$ into query $x$. For a template-augmented rollout, we write $\tilde{x}=\mathcal{T}(x,z)$. The resulting joint objective is:
\begin{equation}
    \label{eq:joint_objective}
    \max_{\theta, \mathcal{Z}} \mathcal{J}_{\text{joint}} = \sum_{x \notin \mathcal{D}_{\text{hard}}} \mathbb{E}_{y \sim \pi_\theta(\cdot \mid x)} [r(x, y)] + \sum_{x \in \mathcal{D}_{\text{hard}}} \max_{z \in \mathcal{Z}} \mathbb{E}_{\tilde y \sim \pi_\theta(\cdot \mid \tilde x)} [r(x, \tilde y)].
\end{equation}
Ordinary rollouts optimize tractable samples, whereas a template supplies an alternative reasoning context for hard samples. Because optimizing discrete templates and continuous policy parameters jointly is highly non-trivial, P$^2$O alternates between policy distillation and prompt evolution.

\subsection{Overview of P$^2$O Framework}

\begin{figure}[t]
  \centering
  \includegraphics[width=1\linewidth]{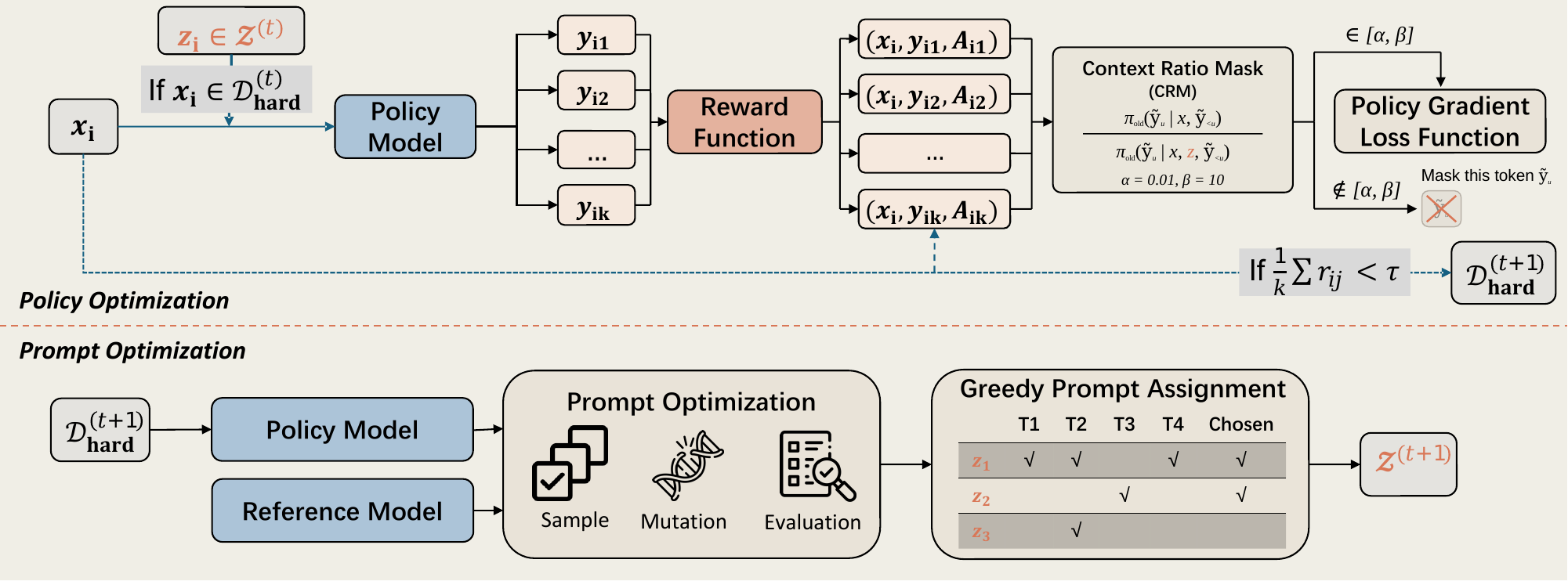}
  \caption{\textbf{Overview of the P$^2$O Framework.} The training process is formulated as an alternating maximization procedure between two phases: (1) Policy Optimization with Context Distillation, where the policy $\pi_\theta$ is updated under a CRM to internalize reasoning patterns elicited by augmented inputs $\tilde{x}$; and (2) Evolutionary Prompt Optimization, where the prompt template set $\mathcal{Z}$ is evolved using GEPA to discover successful trajectories for the remaining hard samples ($\mathcal{D}_{\text{hard}}$).}
  \label{fig:method_main}
\end{figure}

Figure~\ref{fig:method_main} shows how P$^2$O realizes the joint objective. At epoch $t$, we augment the standard GRPO pipeline with three components: (1) \textbf{template-guided rollout}, where an assigned $z\in\mathcal{Z}^{(t)}$ is inserted for each $x\in\mathcal{D}_{\text{hard}}^{(t)}$ to elicit a higher-quality trajectory; (2) \textbf{hard-set update}, where rollout rewards update the policy and determine $\mathcal{D}_{\text{hard}}^{(t+1)}$; and (3) \textbf{prompt evolution}, where GEPA uses this new hard set to produce $\mathcal{Z}^{(t+1)}$ for epoch $t+1$. The complete pseudocode for this alternating loop (Algorithm~\ref{alg:p2o_main}) and the GEPA subroutine (Algorithm~\ref{alg:gepa_sub}) are provided in Appendix~\ref{app:algorithms}.

\paragraph{Threshold-based hard-set definition.}
At epoch $t$, each query $x_i$ receives $K$ rollouts with binary verifiable rewards $r_{ik}\in\{0,1\}$. We define $\tau\in(0,1]$ exclusively as an empirical success-rate threshold:
\begin{equation}
    \label{eq:hard_rate}
    \mathcal{D}_{\text{hard}}^{(t+1)} = \left\{ x_i \in \mathcal{D} \;\middle|\; \widehat p_i^{(t)} \triangleq \frac{1}{K}\sum_{k=1}^K r_{ik} < \tau \right\}.
\end{equation}
In our default configuration, $K=6$ and $\tau=2/6$, so a sample is considered as hard when at most one of its six rollouts succeeds.

\subsection{Phase 1: Policy Optimization with Context Distillation}

We fix $\mathcal{Z}^{(t)}$ while updating $\theta$. For $x\in\mathcal D_{\text{hard}}^{(t)}$, an assigned template $z$ forms $\tilde x=\mathcal T(x,z)$ and elicits a trajectory $\tilde y$; all other queries use bare-context trajectories $y$. The augmented policy is treated as a detached behavior source, while every likelihood used for learning is evaluated under the original query:
\begin{equation}
\label{eq:p2o_gradient}
\begin{aligned}
g_{\text{P}^2\text{O}} \triangleq {}& \sum_{x\notin\mathcal D^{(t)}_{\text{hard}}}\mathbb E_{y\sim\pi_\theta(\cdot\mid x)}\!\left[A(x,y)\nabla_\theta\log\pi_\theta(y\mid x)\right] \\
&+\sum_{x\in\mathcal D^{(t)}_{\text{hard}}}\mathbb E_{\tilde y\sim\textcolor{red}{\operatorname{sg}[\pi_\theta]}(\cdot\mid\textcolor{red}{\tilde x})}\!\left[A(x,\tilde y)\nabla_\theta\log\pi_\theta(\tilde y\mid\textcolor{red}{x})\right].
\end{aligned}
\end{equation}
Here $\operatorname{sg}[\cdot]$ denotes stop-gradient. Decoupling the rollout context $\tilde x$ from the gradient context $x$ distills prompt-elicited reasoning into the policy parameters, following Context Distillation~\citep{snell2022learning}, rather than creating an inference-time prompt dependency. The rewards already collected in this phase also update $\mathcal D_{\text{hard}}^{(t+1)}$ through Eq.~\ref{eq:hard_rate}.

\paragraph{Off-policy correction with Context Ratio Mask.}
The hard-sample trajectory $\tilde{y}$ is generated under the template-augmented context $\tilde{x}$ but optimized under the original query $x$ for context distillation. This introduces a context-induced discrepancy between the rollout and optimization distributions. Inspired by the train--inference discrepancy in  systems~\citep{team2025every}, where excessive distribution deviation can accumulate and destabilize optimization, we introduce a Context Ratio Mask (CRM) to filter out tokens with large likelihood deviations between the two contexts:
\begin{equation}
\label{eq:p2o_objective}
\begin{aligned}
\mathcal{J}_{\mathrm{P}^{2}\mathrm{O}}(\theta)
=\mathbb{E}_{\substack{x\sim\mathcal{D}_{\mathrm{hard}}^{(t)},\\
\tilde y_i\sim\pi_{\mathrm{old}}(\cdot\mid\tilde x_i)}}
\Bigg[\frac{1}{G}\sum_{i=1}^{G}\frac{1}{|\tilde y_i|}\sum_{u=1}^{|\tilde y_i|}
\textcolor{blue}{\mathbf{1}\!\left[\alpha\le
\frac{\pi_{\mathrm{old}}\!\left(\tilde y_{i,u}\mid x,\tilde y_{i,<u}\right)}
     {\pi_{\mathrm{old}}\!\left(\tilde y_{i,u}\mid \tilde x_i,\tilde y_{i,<u}\right)}
\le\beta\right]}\\
\cdot\min\!\Bigg\{
\tilde r_{i,u}\hat A_{i,u},
\operatorname{clip}\!\left(\tilde r_{i,u},1-\epsilon,1+\epsilon\right)\hat A_{i,u}
\Bigg\}\Bigg],
\qquad
\tilde r_{i,u}=
\frac{\pi_{\theta}\!\left(\tilde y_{i,u}\mid x,\tilde y_{i,<u}\right)}
     {\pi_{\mathrm{old}}\!\left(\tilde y_{i,u}\mid x,\tilde y_{i,<u}\right)}.
\end{aligned}
\end{equation}
The context ratio captures the likelihood shift introduced by the auxiliary template. While moderate shifts indicate transferable reasoning signals, excessive deviations suggest template-dependent behaviors that may destabilize training. We therefore set $\alpha=0.01$ and $\beta=10$ to filter out extreme ratios while retaining useful distillation signals.

\subsection{Phase 2: Evolutionary Prompt Optimization}
\label{subsec:phase_2}

After policy optimization, the updated policy $\pi_\theta$ may still fail on a 
subset of challenging instances, resulting in the new hard set 
$\mathcal{D}_{\mathrm{hard}}^{(t+1)}$. These samples indicate regions where 
standard policy exploration remains insufficient. Phase 2 therefore searches for auxiliary prompts that reshape the reasoning context and recover informative 
trajectories for subsequent policy optimization.

With $\pi_\theta$ fixed, we optimize the prompt space by searching for templates 
that maximize the success rate of template-augmented rollouts:

\begin{equation}
    z' \sim \pi_{\text{init}}\left(\cdot \mid \text{prompt}_{\text{meta}}, z, \mathcal{F}_z\right)
\end{equation}
where $\mathcal{F}_z$ contains the query, failed prediction, and verification 
feedback. The generated candidates are evaluated on the hard-sample development 
split and retained through Pareto-based selection. The resulting template set 
$\mathcal{Z}^{(t+1)}$ is then assigned to hard samples for the next epoch.

By alternating template evolution with context distillation, P$^2$O continuously 
expands the set of reachable reasoning trajectories and transfers these gains 
back into the policy.

% Note use of \abovespace and \belowspace to get reasonable spacing
% above and below tabular line

% \subsection{Theorems and Such}
% The preferred way is to number definitions, propositions, lemmas, etc.
% consecutively, within sections, as shown below.
% \begin{definition}
%   \label{def:inj}
%   A function $f:X \to Y$ is injective if for any $x,y\in X$ different, $f(x)\ne
%     f(y)$.
% \end{definition}
% Using \cref{def:inj} we immediate get the following result:
% \begin{proposition}
%   If $f$ is injective mapping a set $X$ to another set $Y$,
%   the cardinality of $Y$ is at least as large as that of $X$
% \end{proposition}
% \begin{proof}
%   Left as an exercise to the reader.
% \end{proof}
% \cref{lem:usefullemma} stated next will prove to be useful.
% \begin{lemma}
%   \label{lem:usefullemma}
%   For any $f:X \to Y$ and $g:Y\to Z$ injective functions, $f \circ g$ is
%   injective.
% \end{lemma}
% \begin{theorem}
%   \label{thm:bigtheorem}
%   If $f:X\to Y$ is bijective, the cardinality of $X$ and $Y$ are the same.
% \end{theorem}
% An easy corollary of \cref{thm:bigtheorem} is the following:
% \begin{corollary}
%   If $f:X\to Y$ is bijective,
%   the cardinality of $X$ is at least as large as that of $Y$.
% \end{corollary}
% \begin{assumption}
%   The set $X$ is finite.
%   \label{ass:xfinite}
% \end{assumption}
% \begin{remark}
%   According to some, it is only the finite case (cf. \cref{ass:xfinite}) that
%   is interesting.
% \end{remark}
%restatable

\section{Experiments}
\label{sec:experiments}

\subsection{Settings}
\label{subsec:settings}

% \paragraph{Dataset Construction.}
% 1. 从deepscaler中随机采样5k条
% 2. 从deepmath中随机采样了5k条，并且过滤了低难度样本，且控制了answer type的比例（也就是控制比如"yes/no"这类数据，但是感觉应该不用写进去？）

% \paragraph{RL Recipe.}
% Following recent practice~\citep{}, we remove the KL loss term. 

% - 训练的通用超参
% 1. KL系数（暂时还没用）
% 2. entropy，没用
% 3. max_prompt_length=4096
% 4. max_response_length=12288
% 5. train_batch_size=128   ppo_mini_batch_size=64
% 6. lr=1e-6
% 7. lr_warmup_steps_ratio=0.285（这个应该不用写？）
% 8. temperature=0.6    rollout.n=6

% - 我们的方法的特有超参
% 1. add_near_hard=True，只错一个的样本也算作hard samples
% 2. auto=heavy_4，大概每次GEPA会产出40个templates（就是通过dev评估且在dev上能涨且加入候选集的templates）
% 3. gepa_select_pareto_k=16，我们的搜索方法是深度优先+广度优先的平衡，即一次搜索16个template（广度优先），然后再进行下一轮搜索（深度优先）
% 4. dev_ratio=300，验证集大小

\paragraph{Datasets.}
We conduct experiments on two datasets: a subset randomly sampled from DeepScaler~\citep{deepscaler2025} and a subset of DeepMath~\citep{he2025deepmath} filtered to difficulty level $\ge 7$. For each dataset, we randomly sample two splits of $N=5{,}000$ and $N=10{,}000$ examples respectively.

\paragraph{Method Configuration.}
We combine GRPO with a one-step off-policy strategy, excluding the KL divergence penalty.
The training hyperparameters include a maximum learning rate of $1 \times 10^{-6}$, a global batch size of 128, and a maximum generation length of 12k tokens.
During the rollout phase, we utilize a temperature of $T=0.6$ and sample $K=6$ trajectories per prompt.
All models are trained for 10 epochs and evaluated with the best dev checkpoint.
We use Qwen3-4B and Qwen3-8B~\citep{yang2025qwen3} as the training backbone.
For GEPA, we use a validation set of 300 examples.
Unlike the original GEPA setting~\citep{agrawal2025gepa}, we use beam size $W$ to parallelize the Pareto search for improved computational efficiency. The candidate selection beam size is set to $W=16$ (yielding $\sim 40$ templates per iteration).
We evaluate two GEPA reflection variants: P$^2$O${_\text{Self-Ref}}$, using the pre-RL model as mutation operator, and P$^2$O${_\text{Teacher-Ref}}$, using Kimi-K2 for reflection.

\paragraph{Reward.}
We employ a strict binary reward ($r \in \{0, 1\}$).
A reward of 1 is granted solely when the response follows the \texttt{\textbackslash boxed\{\}} format and matches the ground truth.

\paragraph{Evaluation Protocol.}
We employ the open-source evaluation suite\footnote{\url{https://github.com/QwenLM/Qwen2.5-Math/tree/main/evaluation}} provided by Qwen for all mathematical benchmarks.
To ensure statistical robustness and encourage exploration on smaller datasets (fewer than 100 samples), we report the average performance across 32 rollouts per question, generated with a temperature of $0.6$.
For larger datasets, we adopt greedy sampling.

\paragraph{Baselines.}
We compare P$^2$O against four baselines throughout the experiments:

(1) \textbf{GRPO}~\citep{shao2024deepseekmath} with varying rollout budgets. This standard RLVR baseline serves to test the naive scaling hypothesis: whether advantage collapse on hard samples can be resolved simply by allocating more compute for exploration.

(2) \textbf{DAPO}~\citep{yu2025dapo}. Our experimental configuration is largely aligned with the original DAPO, incorporating its higher clip parameter, token-level policy-gradient loss, and dynamic sampling strategy; we do not apply DAPO's overlong reward shaping to align with our reward setting. This baseline is included to validate our premise that actively bridging the exploration gap on hard samples yields superior convergence compared to passively discounting them.

(3) \textbf{Single-Turn Reflection}. This ablation replaces GEPA with a single round of self reflection per hard sample at each epoch's end, serving to validate the necessity of GEPA's complex prompt evolution and search strategies.

(4) \textbf{Teacher-Distill-SFT}. A compute-matched baseline designed to test whether P$^2$O$_\text{Teacher-Ref}$'s teacher budget is better spent on data synthesis. It uses an identical teacher budget to generate two rollouts per problem, filters trajectories via ground-truth-verified rejection sampling, and fine-tunes the model via behavioral cloning.

\paragraph{Compute Cost Analysis.}
We account for the overhead of prompt evolution explicitly: the GEPA budget is a hard cap of $10$K rollouts per epoch covering everything in Algorithm~\ref{alg:gepa_sub}, which on DeepMath-5K and DeepScaler-5K amounts to two additional rollouts per training sample per epoch, so GRPO$_{\text{Rollout-8}}$ and P$^2$O with $K=6$ are equivalent in rollout compute.
Table~\ref{tab:rollout_budget} spells out the per-epoch budgets and how each method spends them, including the teacher-side token cost.

% 我们使用Qwen开源的评估套件评估数学benchmark，对于少于100题的数据集，采用温度0.6+32次rollout取平均的方法进行评估，以提高模型的探索性，对于大于100题的数据集，我们采用greedy的方式进行评估，评估长度为16k.

% 就是很简单的format正确（有boxed）+ 答案正确则reward为1，否则为0

\subsection{Main Results}
\label{subsec:main-results}

\begin{table}[t]
\caption{\textbf{Comparative Evaluation on Mathematical Reasoning Benchmarks.} Accuracy (\%) of P$^2$O against the baselines; gains are largest on challenging benchmarks such as AIME24 and AIME25. Best results are highlighted in \textbf{bold}.}
\label{tab:qwen3-math-benchmarks}
\centering
\begin{small}
\begin{sc}
\setlength{\tabcolsep}{4pt}
\resizebox{\textwidth}{!}{%
\begin{tabular}{lccccccc}
\toprule
Model & AIME24 & AIME25 & AMC & MATH500 & Minerva & Olympiad & AVG \\
\midrule
\rowcolor{gray!15} \multicolumn{8}{c}{\textbf{Base Models}} \\\midrule
Qwen3-4B & 23.8 & 19.4 & 68.0 & 82.8 & 31.6 & 49.9 & 45.9 \\
Qwen3-8B & 28.3 & 22.1 & 69.4 & 81.6 & 31.2 & 49.5 & 47.0 \\\midrule
\rowcolor{gray!15} \multicolumn{8}{c}{\textbf{DeepMath-5K}} \\\midrule
Qwen3-4B-GRPO$_{\textnormal{Rollout-6}}$ & 33.8 & 29.0 & 79.5 & 87.6 & 41.5 & 57.5 & 54.8 \\
Qwen3-4B-GRPO$_{\textnormal{Rollout-8}}$ & 35.6 & 28.3 & 79.7 & 89.0 & 40.1 & 56.4 & 54.9 \\
Qwen3-4B-GRPO$_{\textnormal{Rollout-12}}$ & 33.3 & 28.6 & 78.4 & 88.2 & \textbf{43.0} & 56.6 & 54.7 \\
Qwen3-4B-DAPO & 30.0 & 24.4 & 74.0 & 86.8 & 40.8 & 53.5 & 51.6 \\
Qwen3-4B-P$^2$O$_{\textnormal{Single-Turn-Ref}}$ & 27.1 & 22.1 & 74.2 & 85.6 & 41.2 & 52.1 & 50.4 \\
Qwen3-4B-Teacher-Distill-SFT & 28.1 & 25.6 & 73.0 & 85.0 & 36.0 & 52.6 & 50.1 \\
\midrule
\rowcolor{blue!5}
Qwen3-4B-P$^2$O$_{\textnormal{Self-Ref}}$ & \textbf{55.3} & \textbf{40.0} & \textbf{86.2} & \textbf{91.2} & \textbf{43.0} & \textbf{62.4} & \textbf{63.0} \\
\rowcolor{blue!5}
Qwen3-4B-P$^2$O$_{\textnormal{Teacher-Ref}}$ & 46.5 & 32.5 & 82.3 & 88.6 & 41.5 & 59.3 & 58.5 \\
\midrule
Qwen3-8B-GRPO$_{\textnormal{Rollout-6}}$ & 45.6 & 37.1 & 82.8 & 89.6 & 40.4 & 59.9 & 59.2 \\
\midrule
\rowcolor{blue!5}
Qwen3-8B-P$^2$O$_{\textnormal{Self-Ref}}$ & \textbf{57.1} & 41.7 & 86.7 & \textbf{92.2} & \textbf{43.0} & 63.4 & \textbf{64.0} \\
\rowcolor{blue!5}
Qwen3-8B-P$^2$O$_{\textnormal{Teacher-Ref}}$ & 53.4 & \textbf{42.3} & \textbf{88.8} & 91.2 & 40.8 & \textbf{63.9} & 63.4 \\
\midrule
\rowcolor{gray!15} \multicolumn{8}{c}{\textbf{DeepScaler-5K}} \\\midrule
Qwen3-4B-GRPO$_{\textnormal{Rollout-6}}$ & 46.9 & 37.7 & 88.1 & 90.4 & 41.5 & 58.2 & 60.5 \\
Qwen3-4B-GRPO$_{\textnormal{Rollout-8}}$ & 40.6 & 31.0 & 82.5 & 88.0 & 40.4 & 56.1 & 56.4 \\
Qwen3-4B-GRPO$_{\textnormal{Rollout-12}}$ & 41.9 & 32.3 & 80.8 & 86.0 & 43.0 & 57.8 & 57.0 \\
Qwen3-4B-DAPO & 44.4 & 34.2 & 82.0 & 90.0 & 42.6 & 59.7 & 58.8 \\
Qwen3-4B-P$^2$O$_{\textnormal{Single-Turn-Ref}}$ & 25.2 & 22.5 & 70.0 & 82.0 & 36.0 & 51.3 & 47.8 \\
Qwen3-4B-Teacher-Distill-SFT & 28.5 & 24.4 & 70.8 & 81.8 & 32.0 & 49.8 & 47.9 \\
\midrule
\rowcolor{blue!5}
Qwen3-4B-P$^2$O$_{\textnormal{Self-Ref}}$ & \textbf{59.0} & \textbf{42.2} & \textbf{88.4} & \textbf{92.0} & 42.6 & 62.1 & \textbf{64.7} \\
\rowcolor{blue!5}
Qwen3-4B-P$^2$O$_{\textnormal{Teacher-Ref}}$ & 57.7 & 40.2 & 87.0 & 91.6 & \textbf{43.4} & \textbf{64.0} & 64.0 \\
\midrule
Qwen3-8B-GRPO$_{\textnormal{Rollout-6}}$ & 50.8 & 42.9 & 86.4 & 90.8 & 40.1 & 59.0 & 61.7 \\
\midrule
\rowcolor{blue!5}
Qwen3-8B-P$^2$O$_{\textnormal{Self-Ref}}$ & \textbf{61.7} & \textbf{45.4} & \textbf{89.5} & \textbf{91.6} & 41.5 & \textbf{63.6} & \textbf{65.6} \\
\rowcolor{blue!5}
Qwen3-8B-P$^2$O$_{\textnormal{Teacher-Ref}}$ & 55.5 & 45.0 & 85.7 & 91.2 & \textbf{43.0} & 59.6 & 63.3 \\
\midrule
\rowcolor{gray!15} \multicolumn{8}{c}{\textbf{DeepMath-10K}} \\\midrule
Qwen3-4B-GRPO$_{\textnormal{Rollout-6}}$ & 31.9 & 29.2 & 80.2 & 87.6 & 42.6 & 55.7 & 54.5 \\
\midrule
\rowcolor{blue!5}
Qwen3-4B-P$^2$O$_{\textnormal{Self-Ref}}$ & \textbf{48.6} & \textbf{39.2} & 83.0 & 89.8 & 43.4 & 59.6 & \textbf{60.6} \\
\rowcolor{blue!5}
Qwen3-4B-P$^2$O$_{\textnormal{Teacher-Ref}}$ & 46.8 & 33.7 & \textbf{84.8} & \textbf{90.4} & \textbf{43.8} & \textbf{62.5} & 60.3 \\
\midrule
\rowcolor{gray!15} \multicolumn{8}{c}{\textbf{DeepScaler-10K}} \\\midrule
Qwen3-4B-GRPO$_{\textnormal{Rollout-6}}$ & 33.3 & 32.3 & 81.1 & 86.8 & \textbf{41.5} & 55.9 & 55.2 \\
\midrule
\rowcolor{blue!5}
Qwen3-4B-P$^2$O$_{\textnormal{Self-Ref}}$ & \textbf{49.1} & 37.7 & \textbf{85.2} & \textbf{90.4} & 40.8 & \textbf{59.9} & \textbf{60.5} \\
\rowcolor{blue!5}
Qwen3-4B-P$^2$O$_{\textnormal{Teacher-Ref}}$ & 45.3 & \textbf{39.6} & 81.6 & 89.6 & 39.7 & \textbf{59.9} & 59.3 \\
\bottomrule
\end{tabular}%
}
\end{sc}
\end{small}
\end{table}

\begin{figure}[t]
  \centering
  \includegraphics[width=0.95\linewidth]{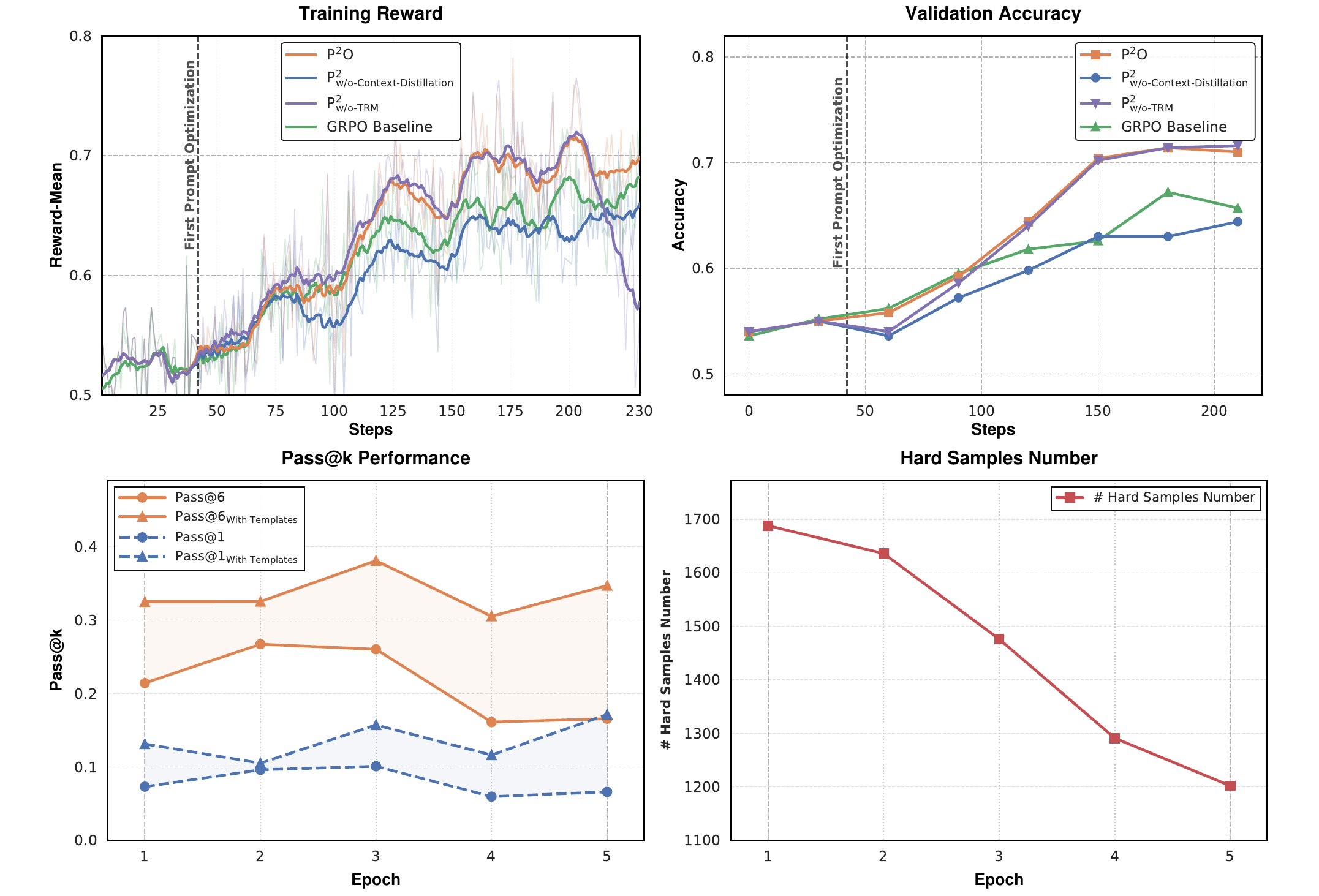}
  \caption{\textbf{Training Dynamics and Prompt Optimization Effectiveness of P$^2$O.} All curves are from the DeepScaler-5K dataset, and the P$^2$O variants use the Teacher-Ref reflection source. \textit{Top:} training reward (left) and validation accuracy (right) of P$^2$O, P$^2$O$_{\textnormal{w/o-CRM}}$, P$^2$O$_{\textnormal{w/o-Context-Distillation}}$, and the GRPO baseline; P$^2$O$_{\textnormal{w/o-CRM}}$ keeps pace for most of training but then collapses, whereas P$^2$O keeps improving, highlighting the necessity of the Context Ratio Mask (Eq.~\ref{eq:p2o_objective}). \textit{Bottom-left:} optimized prompts (triangles) consistently outperform standard prompts (circles) on both Pass@1 and Pass@6. \textit{Bottom-right:} intractable hard samples decline steadily across training epochs.}
  % \caption{\textbf{Training Dynamics of P$^2$O.} Training Reward (Left) and Validation Accuracy (Right) throughout the optimization process using the Teacher-Ref variant on the DeepScaler-5K dataset. P$^2$O maintains a higher reward compared to the baseline by dynamically cracking hard samples. Crucially, this reward advantage translates into robust gains in validation accuracy, confirming that the context distillation mechanism effectively transfers prompt-dependent success into intrinsic model capability.}
  \label{fig:training_dynamics}
\end{figure}

Table~\ref{tab:qwen3-math-benchmarks} summarizes the performance of models trained on DeepMath and DeepScaler across six challenging mathematical benchmarks, comparing P$^2$O against the Qwen3-4B/8B base models and four baselines.

\paragraph{Comparison with GRPO and DAPO.}
As shown in Table~\ref{tab:qwen3-math-benchmarks}, P$^2$O consistently outpaces the policy-optimization baselines by directly addressing the exploration bottleneck on hard samples.
\textbf{P$^2$O breaks the rollout-scaling ceiling under a matched rollout budget}: on DeepMath-5K and DeepScaler-5K, neither a larger GRPO budget nor DAPO's dynamic sampling helps consistently, whereas P$^2$O spends its extra budget on prompt evolution and, with the same $K=6$ rollouts as GRPO$_{\text{Rollout-6}}$, reaches 63.0\% on DeepMath-5K and 64.7\% on DeepScaler-5K ($+8.2$ and $+4.2$ points).
\textbf{Gains persist across data scale and backbones}: on the 10K corpora, P$^2$O leads GRPO$_{\text{Rollout-6}}$ by 6.1 points on DeepMath-10K (60.6\% vs.\ 54.5\%) and by 5.3 points on DeepScaler-10K (60.5\% vs.\ 55.2\%), and on the larger backbone Qwen3-8B-P$^2$O$_{\text{Self-Ref}}$ improves over Qwen3-8B-GRPO$_{\text{Rollout-6}}$ by 4.8 points on DeepMath-5K (64.0\% vs.\ 59.2\%) and by 3.9 points on DeepScaler-5K (65.6\% vs.\ 61.7\%).
\textbf{Gains concentrate on the hardest benchmarks}: on AIME24, P$^2$O$_{\text{Self-Ref}}$ exceeds GRPO$_{\text{Rollout-6}}$ by 21.5 points on DeepMath-5K (55.3\% vs.\ 33.8\%) and by 12.1 points on DeepScaler-5K (59.0\% vs.\ 46.9\%).

\paragraph{Comparison with Reflection and Distillation Baselines.}
P$^2$O also surpasses the two baselines that spend the extra budget on lighter-weight substitutes for GEPA.
Single-Turn Reflection scores only 50.4\% and 47.8\%, below even standard GRPO, and fails to break the difficulty ceiling: the additional gain comes from coupling prompt optimization with policy optimization iteratively, rather than adding either one alone.
Teacher-Distill-SFT spends the same teacher budget on synthesized trajectories and behavioral cloning yet reaches only 50.1\% and 47.9\%, showing that injecting teacher capability as static supervision is far less effective than P$^2$O$_{\text{Teacher-Ref}}$'s context distillation.

\paragraph{Impact of Reflection Models on Prompt Evolution.}
P$^2$O$_{\text{Self-Ref}}$ outperforms P$^2$O$_{\text{Teacher-Ref}}$ in all four settings, with the advantage most pronounced on DeepMath-5K (63.0\% vs.\ 58.5\% avg.), where the two variants differ by $4.5$ points, and gains of $0.7$, $0.3$, and $1.2$ points on DeepScaler-5K, DeepMath-10K, and DeepScaler-10K, respectively.
This indicates that a stronger external teacher does not necessarily yield a more useful mutation operator: self-generated refinements, which stay close to the policy's own reasoning capacity,
are easier to internalize, whereas teacher priors may be harder to absorb.
We examine this inversion in detail in Appendix~\ref{app:template_quality}, where the stronger teacher evolves templates with higher dev scores yet yields a substantially weaker model.
Crucially, both variants consistently surpass the GRPO baseline across all datasets,
confirming that GEPA-driven structural exploration robustly improves performance regardless of the reflection source.

% \paragraph{Scaling to Larger Datasets.}
% To assess whether P$^2$O remains effective as the training data scales up, we further evaluate on the 10K variants of both datasets. As shown in Table~\ref{tab:qwen3-math-benchmarks}, P$^2$O continues to outperform the GRPO baseline on both DeepMath-10K and DeepScaler-10K. Specifically, P$^2$O$_{\text{Self-Ref}}$ achieves 60.6\% on DeepMath-10K and 60.5\% on DeepScaler-10K, surpassing GRPO$_{\text{Rollout-6}}$ by 6.1\% and 5.3\%, respectively, demonstrating that the benefits of prompt-guided exploration do not diminish with larger training corpora.

\subsection{Training Dynamics}

As shown in Figure~\ref{fig:training_dynamics}~(top), the continuous integration of optimized prompts maintains a training reward consistently superior to that of the GRPO baseline.
By contrast, P$^2$O$_{\textnormal{w/o-CRM}}$ tracks P$^2$O for most of the optimization and then collapses abruptly in the late stage, corroborating the design of the Context Ratio Mask in Eq.~\ref{eq:p2o_objective}: without the mask, the rare but extreme template-induced likelihood ratios accumulate into unstable updates.
Crucially, this advantage of P$^2$O over the GRPO baseline translates into gains in validation accuracy, confirming that the model effectively internalizes the elicited reasoning patterns to achieve robust in-distribution generalization. 

Meanwhile, as shown in Figure~\ref{fig:training_dynamics}~(bottom), as the model's intrinsic capability grows during training, GEPA effectively assists in continuously conquering ``hard samples'', as evidenced by the steady decline in the number of intractable instances (Right).
Furthermore, the performance breakdown (Left) reveals that prompt optimization yields consistent gains across both Pass@1 and Pass@6 metrics.
This demonstrates that the evolved templates do not merely facilitate a single lucky guess; rather, they robustly enhance the model's solution space, boosting both the deterministic success rate and the broader exploration coverage required for effective distillation.
Moreover, introducing prompt optimization into policy optimization does indeed alleviate advantage collapse (Appendix~\ref{app:advantage_collapse}).

\subsection{Ablations}
\label{subsec:ablations}

\begin{table}[t]
\caption{\textbf{Ablation Study of P$^2$O Components.} Comparison of performance on mathematical benchmarks when excluding context distillation or group prompt diversity. Data points are derived from the Teacher-Ref w/o-CRM variant on the DeepScaler-5K dataset.}
\label{tab:ablation_study}
\centering 
\begin{small}
\begin{sc}
\setlength{\tabcolsep}{4pt}
\resizebox{\textwidth}{!}{%
\begin{tabular}{lccccccc}
\toprule
Model & AIME24 & AIME25 & AMC & MATH500 & Minerva & Olympiad & AVG \\
\midrule
Qwen3-4B-GRPO & 46.9 & 37.7 & 88.1 & 90.4 & \textbf{41.5} & 58.2 & 60.5 \\
Qwen3-4B-P$^2$O & \textbf{59.8} & \textbf{49.4} & \textbf{92.2} & \textbf{91.4} & 36.4 & \textbf{62.2} & \textbf{65.2} \\\midrule
Qwen3-4B-P$^2$O$_{\textnormal{w/o-Context-Distillation}}$ & 36.5 & 31.5 & 81.1 & 89.6 & 40.1 & 54.8 & 55.6 \\
Qwen3-4B-P$^2$O$_{\textnormal{Same-Template-in-Group}}$ & 57.3 & 44.6 & 88.9 & 90.8 & 40.4 & 63.0 & 64.2 \\
\bottomrule
\end{tabular}%
}
\end{sc}
\end{small}
\end{table}

\paragraph{Importance of Context Distillation.}
We compare Qwen3-4B-P$^2$O against its w/o-Context-Distillation variant, which differs only in how template-augmented trajectories contribute gradients (Section~\ref{app:context_distillation}). As shown in Table~\ref{tab:ablation_study}, this causes a severe drop: 65.2\%$\to$55.6\%, even 4.9\% below the GRPO baseline (60.5\%).
Training on $(\tilde{x}, \tilde y)$ induces a ``dependency'' effect, where the model relies on the auxiliary prompt $z$ rather than internalizing the reasoning logic, and thus fails to generalize when only $x$ is available at evaluation.
Context distillation is therefore a prerequisite, not merely an enhancement, for transferring the teacher reference's capabilities into the model's intrinsic parameters.

\paragraph{Impact of Group Prompt Diversity.}
Standard P$^2$O induces template diversity across the $K$ rollouts per query via the greedy assignment (Algorithm~\ref{alg:greedy_assign}), whereas P$^2$O$_\text{Same-Template-in-Group}$ fixes one sampled template $z \sim \mathcal{Z}^{(t)}$ for all $K$ rollouts. As shown in Table~\ref{tab:ablation_study}, the diversity-driven approach achieves 65.2\% average accuracy, outperforming the single-template baseline (64.2\%), with pronounced gains on AIME24 (+2.5\%) and AIME25 (+4.8\%). This confirms that diverse Pareto-optimal prompts within a group broaden reasoning space coverage, yielding more diverse supervision signals and more effective distillation of complex reasoning capabilities.

% ---------------------------------------------------------------------------
% Both tables share one float and are placed side by side, each in a minipage
% with \minitablecaption{...} so that they are numbered separately.
% $\Delta$ entries use \gain{...} (green up arrow) and \drop{...} (red down arrow).
% ---------------------------------------------------------------------------
\begin{table}[t]
\centering
\begin{minipage}[t]{0.48\textwidth}
\centering
\minitablecaption{\textbf{Effect of the Context Ratio Mask.} Six-benchmark average accuracy (\%) per training set, backbone, and reflection source; $\Delta$ (P$^2$O $-$ w/o-CRM) marked \textcolor{green!55!black}{$\uparrow$}/\textcolor{red}{$\downarrow$}. See Table~\ref{tab:CRM_ablation_full}.}
\vspace{0.5em}
\label{tab:CRM_ablation}
\footnotesize
\setlength{\tabcolsep}{4pt}
\begin{tabular}{@{}lccc@{}}
\toprule
Variant & P$^2$O & w/o-CRM & $\Delta$ \\
\midrule
\multicolumn{4}{@{}l}{\textbf{DeepMath-5K}} \\
\hspace{0.8em}Qwen3-4B Self-Ref & 63.0 & 64.2 & \drop{1.2} \\
\hspace{0.8em}Qwen3-4B Teacher-Ref & 58.5 & 57.9 & \gain{0.6} \\
\hspace{0.8em}Qwen3-8B Self-Ref & 64.0 & 61.3 & \gain{2.7} \\
\hspace{0.8em}Qwen3-8B Teacher-Ref & 63.4 & 61.0 & \gain{2.4} \\
\midrule
\multicolumn{4}{@{}l}{\textbf{DeepScaler-5K}} \\
\hspace{0.8em}Qwen3-4B Self-Ref & 64.7 & 62.4 & \gain{2.3} \\
\hspace{0.8em}Qwen3-4B Teacher-Ref & 64.0 & 65.2 & \drop{1.2} \\
\hspace{0.8em}Qwen3-8B Self-Ref & 65.6 & 62.2 & \gain{3.4} \\
\hspace{0.8em}Qwen3-8B Teacher-Ref & 63.3 & 60.6 & \gain{2.7} \\
\bottomrule
\end{tabular}
\end{minipage}\hfill
\begin{minipage}[t]{0.48\textwidth}
\centering
\minitablecaption{\textbf{Hyper-parameter Ablation of the Context Ratio Mask.} Six-benchmark average accuracy (\%) for Qwen3-8B on the DeepMath-5K training set with Self-Ref, as the mask bounds $\alpha$ and $\beta$ vary; ``w/o CRM'' removes the mask entirely. The default interval $[0.01,10]$ reaches 64.0\%, 2.7 points above w/o CRM. Best result in \textbf{bold}.}
\vspace{0.6em}
\label{tab:CRM_hyperparam}
\footnotesize
\setlength{\tabcolsep}{6pt}
\begin{tabular}{@{}cccc@{}}
\toprule
$\alpha$ & $\beta$ & Qwen3-8B & $\Delta$ vs. w/o CRM \\
\midrule
\multicolumn{2}{c}{w/o CRM} & 61.3 & -- \\
\midrule
0.8 & 1.2 & 58.3 & \drop{3.0} \\
0.05 & 5 & 63.1 & \gain{1.8} \\
0.01 & 10 & \textbf{64.0} & \textbf{\gain{2.7}} \\
0.005 & 50 & 61.5 & \gain{0.2} \\
\bottomrule
\end{tabular}
\end{minipage}
\end{table}

\paragraph{Effect of the Context Ratio Mask.}
The Context Ratio Mask of Eq.~\ref{eq:p2o_objective} discards the tokens whose likelihood departs too far from the original-query distribution under the template context. Table~\ref{tab:CRM_ablation} compares P$^2$O with its w/o-CRM counterpart, which differs only in this mask, across both training sets, backbones, and reflection sources, all with the default interval $[\alpha,\beta]=[0.01,10]$. The benefit of the mask grows with the scale of the backbone: with Qwen3-8B, removing it costs 2.4--3.4 points in all four settings and the loss concentrates on the hardest benchmarks (AIME24 drops by 4.2--8.6 points, AIME25 by 2.3--7.1), whereas with Qwen3-4B the effect is mixed, ranging from $-1.2$ to $+2.3$ and averaging $+0.1$, for a net gain of $+1.5$ points across the eight comparisons. Filtering template-dominated tokens thus pays off only when the policy is strong enough to exploit the remaining transferable signal, so we keep the mask in every main configuration.

We also sweep the bounds $\alpha$ and $\beta$ (Table~\ref{tab:CRM_hyperparam}). The sweep is single-peaked: tightening the interval discards the template-elicited tokens carry the transferable gain, loosening it admits the extreme ratios behind the late-stage collapse of Figure~\ref{fig:training_dynamics}, and our default $[0.01,10]$ attains the best accuracy.

\section{Related Works}

% 先谈GRPO和RLVR，然后说GRPO和RLVR的缺陷，然后引出DAPO和另一个工作。然后说现在的hint-based方法很多依赖于已有的ground truth answer fragments，没有"optimization"的意味在里面.

RLVR and GRPO are widely used for LLM reasoning alignment, stabilizing training via group baselines and verification, but still struggle with exploration in sparse or misleading landscapes.
To address these scalability and stability issues, recent works have focused on algorithmic advances. 
For instance, DAPO \citep{yu2025dapo} prunes prompts with accuracy equal to $1$ or $0$ to improve training stability. 
Despite these improvements, these methods primarily focus on optimizing the policy on a fixed search space, leaving the initial task prompts untouched.

To further aid the model in traversing complex reasoning paths, various ``hint-based'' strategies have been proposed, including strong model guidance~\citep{nath2025adaptive, liu2025ghpo}, experience replay data~\citep{zhan2025exgrpo}, and expert solutions~\citep{zhang2025bread, yan2025learning, li2025questa}, which attempt to scaffold the reasoning process.
Similarly, Critique-GRPO~\citep{zhang2025critique} utilizes feedback signals to guide the policy.
However, most of these methods rely heavily on external expert supervision or expert trajectory fragments. In contrast, P$^2$O achieves self-improvement without external guidance. By treating the prompt as an optimizable parameter and evolving it jointly with the policy, P$^2$O matches or exceeds the performance of teacher-dependent variants while remaining compatible with teacher reflection.

\section{Conclusion and Limitations}
\label{sec:conclusion}

In this paper, we introduce P$^2$O, a novel framework that bridges the gap between discrete prompt evolution and continuous policy optimization to overcome the collapse of advantages in RLVR. By dynamically identifying hard samples and leveraging the GEPA algorithm, P$^2$O evolves targeted prompt templates that guide the policy toward discovering successful reasoning trajectories that are otherwise inaccessible. Crucially, our context distillation mechanism ensures that these prompt-elicited capabilities are internalized directly into the model parameters, eliminating reliance on inference-time guidance. Extensive evaluations across challenging mathematical reasoning benchmarks demonstrate that P$^2$O significantly outperforms standard GRPO paradigms, establishing joint optimization as a robust pathway for autonomous self-improvement in LLMs.

% ---------------------------------------------------------------------------
% AI use statement: required by ICLR 2027, must not exceed 1 page, and does not
% count toward the page limit. The heading matches the boilerplate of the
% official template (iclr2027_conference_backup.tex); see also
% https://iclr.cc/Conferences/2027/AIPolicyForAuthors
% ---------------------------------------------------------------------------
\subsection*{AI use statement}

In this work, we used generative AI tools for two tasks with required disclosure: implementing methods and formulating mathematical claims. Our implementation modifies the open-source \textsc{verl} codebase, where AI assistance was used to write parts of the core training, evaluation, and test code---reviewed and tested by the two authors who led the implementation---and to formulate and refine the mathematical claims and theoretical presentation of our method (Section~\ref{sec:method}). We have not used generative AI tools for the other required-disclosure tasks: the research questions and hypotheses, research methodology, experimental design, and interpretation of results were produced by the authors, and the data are randomly sampled from public open-source corpora; the remaining tasks are not applicable to this work, which contains no formal proofs and no surveys or interviews. Additionally, we used generative AI tools for tasks with recommended disclosure: creating and editing software code and drafting and editing parts of this manuscript, including rewriting and condensing passages, reformulating table captions and layout, and checking the internal consistency of the reported numbers.

We have reviewed all AI-assisted work: the assisted code was verified and tested, the assisted mathematical statements were re-derived by the authors, and all AI-suggested edits were checked against our experimental records and the original sources. We take responsibility for the final content of this work, including text, claims, or artifacts produced with the aid of generative AI.

% ---------------------------------------------------------------------------
% Ethics and reproducibility statements: recommended by ICLR 2027 and not
% counted toward the page limit (see the official template,
% iclr2027_conference_backup.tex).
% ---------------------------------------------------------------------------
\subsection*{Ethics statement}

This work studies policy and prompt optimization for mathematical reasoning with verifiable rewards. All datasets, backbone models, and evaluation suites we use are publicly available, and the benchmarks are standard, publicly released mathematics problem sets that contain no personally identifiable information. The study involves no human subjects and no private or sensitive data, and we identify no ethical concerns specific to this work.

\subsection*{Reproducibility statement}

The complete implementation, including the training pipeline, the GEPA-based prompt evolution, the evaluation scripts, and the configuration files used to produce all results, is submitted as anonymous supplementary material. The datasets, backbone models, optimization hyper-parameters, and GEPA settings are specified in Section~\ref{subsec:settings}; pseudocode for the P$^2$O outer loop, the greedy template assignment, and the GEPA subroutines is given in Appendix~\ref{app:algorithms}, and the context distillation objective in Appendix~\ref{app:context_distillation}. All benchmarks are evaluated with the open-source Qwen evaluation suite, and our main comparisons against the baselines are reported in Table~\ref{tab:qwen3-math-benchmarks}.

\bibliography{example_paper}
\bibliographystyle{iclr2027_conference}

\clearpage
\appendix
\section{Algorithmic Details}
\label{app:algorithms}

This appendix gives the pseudocode for the P$^2$O outer loop, sample-specific template assignment, and the GEPA subroutines used in Phase 2. The first two algorithms describe the interface introduced by P$^2$O; the latter two provide the implementation details of GEPA.

\begin{algorithm}[t]
\small
   \caption{P$^2$O Training Loop}
   \label{alg:p2o_main}
\begin{algorithmic}[1]
   \REQUIRE Training data $\mathcal{D}$, policy $\pi_\theta$, reference model $\pi_{\text{init}}$
   \REQUIRE Epochs $N_{\text{epoch}}$, rollout number $K$, threshold $\tau$
   \ENSURE Optimized policy $\pi_{\theta^*}$

   \STATE Initialize $\mathcal{D}_{\text{hard}}^{(0)}\gets\emptyset$ and $\mathcal{Z}^{(0)}\gets\emptyset$
   \FOR{$t=0$ \textbf{to} $N_{\text{epoch}}-1$}
      \STATE Initialize $\mathcal{D}_{\text{hard}}^{(t+1)}\gets\emptyset$
      \FORALL{$(x_i,\hat y_i)\in\mathcal{D}$}
         \FOR{$k=1$ \textbf{to} $K$}
            \IF{$x_i\in\mathcal{D}_{\text{hard}}^{(t)}$}
               \STATE Sample $z\sim\mathcal{Z}^{(t)}$ and roll out $y^{\mathrm{roll}}_{ik}\sim\pi_\theta(\cdot\mid\mathcal{T}(x_i,z))$
            \ELSE
               \STATE Roll out $y^{\mathrm{roll}}_{ik}\sim\pi_\theta(\cdot\mid x_i)$
            \ENDIF
            \STATE $r_{ik}\gets\textsc{RewardFunc}(y^{\mathrm{roll}}_{ik},\hat y_i)$
         \ENDFOR
         \STATE $\widehat p_i^{(t)}\gets K^{-1}\sum_{k=1}^{K}r_{ik}$
         \IF{$\widehat p_i^{(t)}<\tau$}
            \STATE $\mathcal{D}_{\text{hard}}^{(t+1)}\gets\mathcal{D}_{\text{hard}}^{(t+1)}\cup\{x_i\}$
         \ENDIF
         \STATE Compute $A_{ik}$ and update $\pi_\theta$ on the original $x_i$ using Eqs.~\ref{eq:p2o_gradient} and~\ref{eq:p2o_objective}
      \ENDFOR
      \IF{$\mathcal{D}_{\text{hard}}^{(t+1)}\neq\emptyset$}
         \STATE $\mathcal{Z}^{(t+1)}\gets\textsc{Gepa}(\mathcal{D}_{\text{hard}}^{(t+1)},\pi_\theta,\pi_{\text{init}})$
         \STATE Assign templates for the next epoch with \textsc{GreedyPromptAssignment}
      \ELSE
         \STATE $\mathcal{Z}^{(t+1)}\gets\emptyset$
      \ENDIF
   \ENDFOR
   \STATE \textbf{return} $\pi_{\theta^*}$
\end{algorithmic}
\end{algorithm}

\begin{algorithm}[t]
\small
   \caption{\textsc{GreedyPromptAssignment}}
   \label{alg:greedy_assign}
\begin{algorithmic}[1]
   \REQUIRE Candidate templates and dev scores $\mathcal{Z}=\{(z_i,\mathbf r_i)\}_{i=1}^{M}$
   \REQUIRE Hard samples $\mathcal{D}_{\text{hard}}$, rollout number $K$
   \ENSURE Sample-specific assignments $\mathcal{A}$
   \STATE Let $\mathcal{C}(z_i)=\{x_j:r_{i,j}=1\}$ denote the dev samples solved by $z_i$
   \STATE Initialize covered templates $\mathcal{Z}_{\text{covered}}\gets\emptyset$ and samples $\mathcal{S}_{\text{covered}}\gets\emptyset$
   \WHILE{$\exists z_i:\mathcal{C}(z_i)\setminus\mathcal{S}_{\text{covered}}\neq\emptyset$}
      \STATE Select $z^*\gets\arg\max_{z_i\notin\mathcal{Z}_{\text{covered}}}|\mathcal{C}(z_i)\setminus\mathcal{S}_{\text{covered}}|$
      \STATE Add $z^*$ to $\mathcal{Z}_{\text{covered}}$ and update $\mathcal{S}_{\text{covered}}\gets\mathcal{S}_{\text{covered}}\cup\mathcal{C}(z^*)$
   \ENDWHILE
   \STATE Compute $\bar r_i\gets N^{-1}\sum_{n=1}^{N}r_{i,n}$ as the sampling weight of each covered template
   \STATE Initialize $\mathcal{A}\gets\emptyset$
   \FORALL{$x_j\in\mathcal{D}_{\text{hard}}$}
      \IF{$x_j\in\mathcal{S}_{\text{covered}}$}
         \STATE $\mathcal{Z}_j\gets\{z_i\in\mathcal{Z}_{\text{covered}}:x_j\in\mathcal{C}(z_i)\}$
      \ELSE
         \STATE $\mathcal{Z}_j\gets\mathcal{Z}_{\text{covered}}$
      \ENDIF
      \STATE Sample $\mathbf z_j=(z_{j,1},\ldots,z_{j,K})$ from $\mathcal{Z}_j$ weighted by $\{\bar r_i\}$ and add $(x_j,\mathbf z_j)$ to $\mathcal{A}$
   \ENDFOR
   \STATE \textbf{return} $\mathcal{A}$
\end{algorithmic}
\end{algorithm}

\begin{algorithm}[t]
   \caption{\textsc{Gepa}}
   \label{alg:gepa_sub}
\begin{algorithmic}[1]
   \REQUIRE Hard Data $\mathcal{D}_{\text{hard}}$, Policy Model $\pi_{\theta}$, Reference Model $\pi_{\text{init}}$
   \REQUIRE Budget $C_{\text{total}}$, Mini-batch $B$, Width $W$
   \ENSURE Output Template Set $\mathcal{Z}$
   
   \STATE Split $\mathcal{D}_{\text{hard}}$ into $\mathcal{D}_{\text{hard}}^{\text{train}}$ and $\mathcal{D}_{\text{hard}}^{\text{dev}}$
   \STATE Initialize $\mathcal{Z}$ with the empty template evaluated on $\mathcal{D}_{\text{hard}}^{\text{dev}}$
   \STATE $C_{\text{left}} \gets C_{\text{total}}$
   
   \WHILE{$C_{\text{left}} > 0$}
      \STATE $\mathcal{Z}_{\text{front}} \gets \textsc{SelectParetoFront}(\mathcal{Z}, W)$
      
      \FORALL{$z \in \mathcal{Z}_{\text{front}}$}
         \STATE Sample mini-batch $\mathcal{D}_{\text{mini}} \subset \mathcal{D}_{\text{hard}}^{\text{train}}$ of size $B$
         \STATE $\bar{r}_{\text{old}} \gets \operatorname{mean}(\textsc{Eval}(\pi_{\theta}, z, \mathcal{D}_{\text{mini}}))$
         
         \STATE Generate error feedback $\mathcal{F}$ from rollouts and rewards on $\mathcal{D}_{\text{mini}}$
         \STATE $z' \gets \pi_{\text{init}}(\text{``Propose Improvement''}, z, \mathcal{F})$
         \STATE $\bar{r}_{\text{new}} \gets \operatorname{mean}(\textsc{Eval}(\pi_{\theta}, z', \mathcal{D}_{\text{mini}}))$
         
         \STATE Decrease $C_{\text{left}}$ by the two mini-batch evaluations
         
         \IF{$\bar{r}_{\text{new}} > \bar{r}_{\text{old}}$}
            \STATE $\mathcal{Z} \gets \mathcal{Z} \cup \{(z', \textsc{Eval}(\pi_{\theta}, z',\mathcal{D}_{\text{hard}}^{\text{dev}}))\}$
            \STATE Decrease $C_{\text{left}}$ by the development evaluation cost
         \ENDIF
      \ENDFOR
   \ENDWHILE

   \STATE \textbf{return} $\mathcal{Z}$
\end{algorithmic}
\end{algorithm}

\begin{algorithm}[t]
   \caption{\textsc{SelectParetoFront}}
   \label{alg:select_pareto}
\begin{algorithmic}[1]
   \REQUIRE Template set with scores $\mathcal{Z} = \{(z_1, (r_{1,1}, \ldots, r_{1,N})), \ldots, (z_M, (r_{M,1}, \ldots, r_{M,N}))\}$, Width $W$
   \ENSURE Selected templates $\mathcal{Z}_{\text{front}}$
   \STATE A template $z_j$ dominates $z_i$ if it is no worse on every dev sample and better on at least one
   \STATE $\mathcal{Z}_{\text{front}} \gets \emptyset$
   \FOR{$i = 1$ \textbf{to} $M$}
      \STATE $\text{dominated} \gets \textsc{False}$
      \FOR{$j = 1$ \textbf{to} $M$}
         \IF{$j \neq i$ \textbf{and} $z_j$ dominates $z_i$}
            \STATE $\text{dominated} \gets \textsc{True}$
            \STATE \textbf{break}
         \ENDIF
      \ENDFOR
      \IF{\textbf{not} dominated}
         \STATE $\mathcal{Z}_{\text{front}} \gets \mathcal{Z}_{\text{front}} \cup \{(z_i, (r_{i,1}, \ldots, r_{i,N}))\}$
      \ENDIF
   \ENDFOR
   \IF{$|\mathcal{Z}_{\text{front}}| \leq W$}
      \STATE $\mathcal{Z}_{\text{front}} \gets \{z \mid (z, (r_{1}, \ldots, r_{N})) \in \mathcal{Z}_{\text{front}}\}$
   \ELSE
      \STATE Compute weights: $\bar{r}_i \gets \frac{1}{N}\sum_{n=1}^{N} r_{i,n}$ for each $(z_i, (r_{i,1}, \ldots, r_{i,N})) \in \mathcal{Z}_{\text{front}}$
      \STATE Sample $W$ templates from $\mathcal{Z}_{\text{front}}$ weighted by $\{\bar{r}_i\}$
      \STATE $\mathcal{Z}_{\text{front}} \gets \{\text{sampled templates}\}$
   \ENDIF
   \STATE \textbf{return} $\mathcal{Z}_{\text{front}}$
\end{algorithmic}
\end{algorithm}

\clearpage
\section{Per-Epoch Rollout Budget}
\label{app:rollout_budget}

Table~\ref{tab:rollout_budget} spells out how each method spends its rollouts within one epoch. The GEPA budget is a hard cap: prompt evolution may consume at most $10$K rollouts per epoch, independent of the number of search iterations, and this cap covers everything charged in Algorithm~\ref{alg:gepa_sub}---mini-batch rollouts, reflection calls (including all Kimi-K2 calls), and dev evaluations. On the DeepMath-5K and DeepScaler-5K training sets ($N=5{,}000$), the cap is exactly two extra rollouts per training sample, which is why GRPO$_{\text{Rollout-8}}$ and P$^2$O with $K=6$ are compute-equivalent in terms of rollouts.
The two rollout-matched budgets are nevertheless spent in different places: most of P$^2$O's rollouts are mini-batch rollouts and dev evaluations produced by the training policy itself, and the teacher is invoked only for reflection, so P$^2$O$_{\text{Teacher-Ref}}$ has a total teacher-side token cost below $0.5$M tokens, whereas Teacher-Distill-SFT spends about $14$M teacher tokens to synthesize its supervision data.

\begin{table}[h]
\caption{\textbf{Per-Epoch Rollout Budgets.} All entries are rollout counts per epoch, where $N$ is the number of training samples ($N=5{,}000$ for DeepMath-5K and DeepScaler-5K, so that $2N=10$K). \emph{GEPA rollouts} include everything charged to the budget of Algorithm~\ref{alg:gepa_sub}: mini-batch rollouts, reflection calls (including all Kimi-K2 calls), and all dev evaluations; GRPO uses none. Because the GEPA budget is a hard cap, P$^2$O spends $8N$ rollouts per epoch at $K=6$ and therefore matches GRPO$_{\text{Rollout-8}}$.}
\label{tab:rollout_budget}
\centering
\begin{small}
\begin{sc}
\setlength{\tabcolsep}{8pt}
\begin{tabular}{lccc}
\toprule
Method & Policy Rollouts & GEPA Rollouts & Total Rollouts/Epoch \\
\midrule
GRPO$_{\textnormal{Rollout-6}}$ & $6N$ & -- & $6N$ \\
GRPO$_{\textnormal{Rollout-8}}$ & $8N$ & -- & $8N$ \\
GRPO$_{\textnormal{Rollout-12}}$ & $12N$ & -- & $12N$ \\
\midrule
P$^2$O$_{\textnormal{Self-Ref}}$ ($K=6$) & $6N$ & $2N$ & $8N$ \\
P$^2$O$_{\textnormal{Teacher-Ref}}$ ($K=6$) & $6N$ & $2N$ & $8N$ \\
\bottomrule
\end{tabular}
\end{sc}
\end{small}
\end{table}

\clearpage
\section{Context Distillation Objective}
\label{app:context_distillation}

We compare two variants that use identical bare-context rollouts and the same template-augmented trajectories for hard samples. They differ only in whether the augmented policy acts as a detached behavior source and whether the resulting trajectory is optimized under the original or augmented context. The differing terms are highlighted in red below.

\paragraph{P$^2$O with context distillation (P$^2$O).}
Our configuration reuses Eq.~\ref{eq:p2o_gradient}:
\begin{equation}
\begin{aligned}
g_{\text{P}^2\text{O}} \triangleq {}& \sum_{x\notin\mathcal D^{(t)}_{\text{hard}}}\mathbb E_{y\sim\pi_\theta(\cdot\mid x)}\!\left[A(x,y)\nabla_\theta\log\pi_\theta(y\mid x)\right] \\
&+\sum_{x\in\mathcal D^{(t)}_{\text{hard}}}\mathbb E_{\tilde y\sim\textcolor{red}{\operatorname{sg}[\pi_\theta]}(\cdot\mid\tilde x)}\!\left[A(x,\tilde y)\nabla_\theta\log\pi_\theta(\tilde y\mid\textcolor{red}{x})\right].
\end{aligned}
\label{eq:appendix_context_distillation}
\end{equation}
The stop-gradient policy generates a template-elicited trajectory from $\tilde x$, but the likelihood term is evaluated under the original query $x$. Consequently, the update transfers the elicited behavior into $\pi_\theta(\cdot\mid x)$.

\paragraph{P$^2$O without context distillation (P$^2$O$_{\textnormal{w/o-Context-Distillation}}$).}
The ablated variant instead uses
\begin{equation}
\begin{aligned}
g_{\text{w/o-Context-Distillation}} \triangleq {}& \sum_{x\notin\mathcal D^{(t)}_{\text{hard}}}\mathbb E_{y\sim\pi_\theta(\cdot\mid x)}\!\left[A(x,y)\nabla_\theta\log\pi_\theta(y\mid x)\right] \\
&+\sum_{x\in\mathcal D^{(t)}_{\text{hard}}}\mathbb E_{\tilde y\sim\textcolor{red}{\pi_\theta}(\cdot\mid\tilde x)}\!\left[A(x,\tilde y)\nabla_\theta\log\pi_\theta(\tilde y\mid\textcolor{red}{\tilde x})\right].
\end{aligned}
\label{eq:appendix_no_context_distillation}
\end{equation}
Here, the augmented policy is optimized in the same context from which it samples. This update improves $\pi_\theta(\cdot\mid\tilde x)$ but does not explicitly transfer the prompt-induced behavior to $\pi_\theta(\cdot\mid x)$, allowing the model to remain dependent on the auxiliary template at evaluation time.

\section{Training Dynamics of GRPO}
\label{app:grpo_dynamics}

\begin{figure}[h]
  \centering
  \includegraphics[width=1\linewidth]{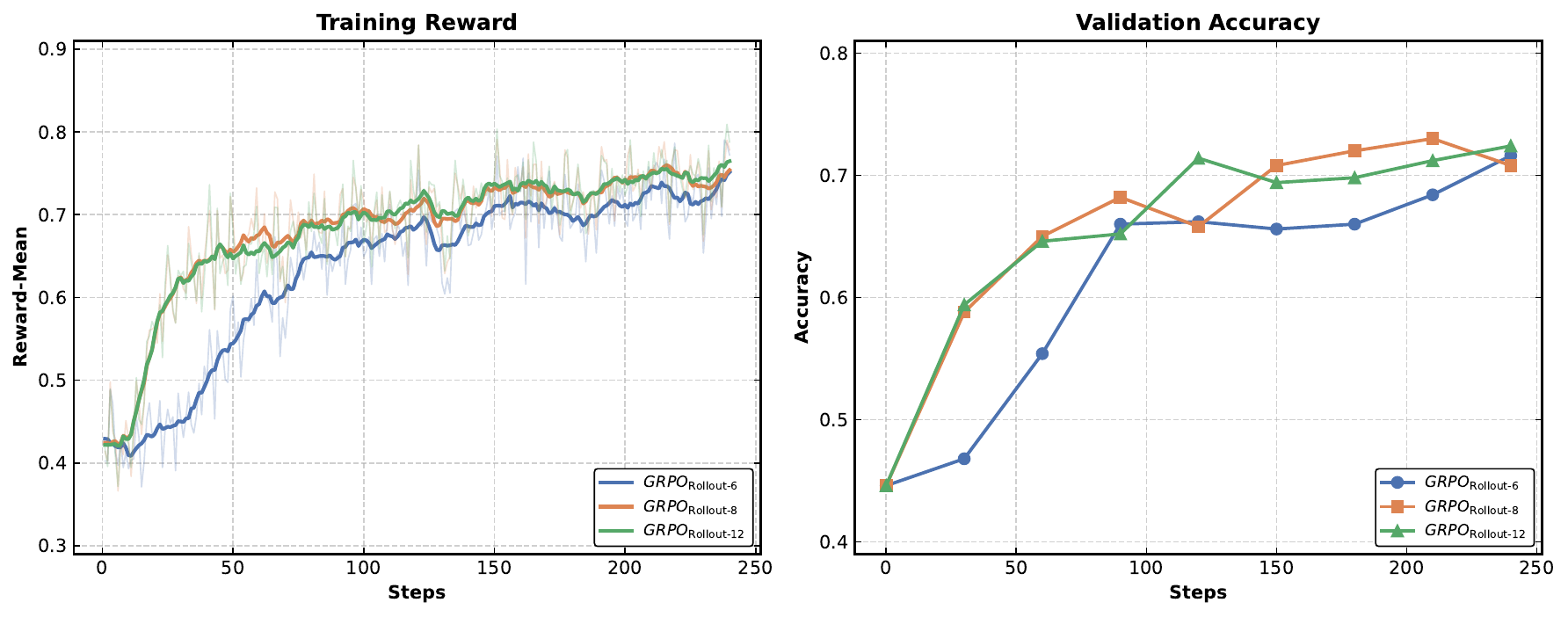}
  \caption{Training Dynamics of GRPO on DeepMath-5K dataset}
  \label{fig:training_dynamics_deepmath}
\end{figure}

\begin{figure}[h]
  \centering
  \includegraphics[width=1\linewidth]{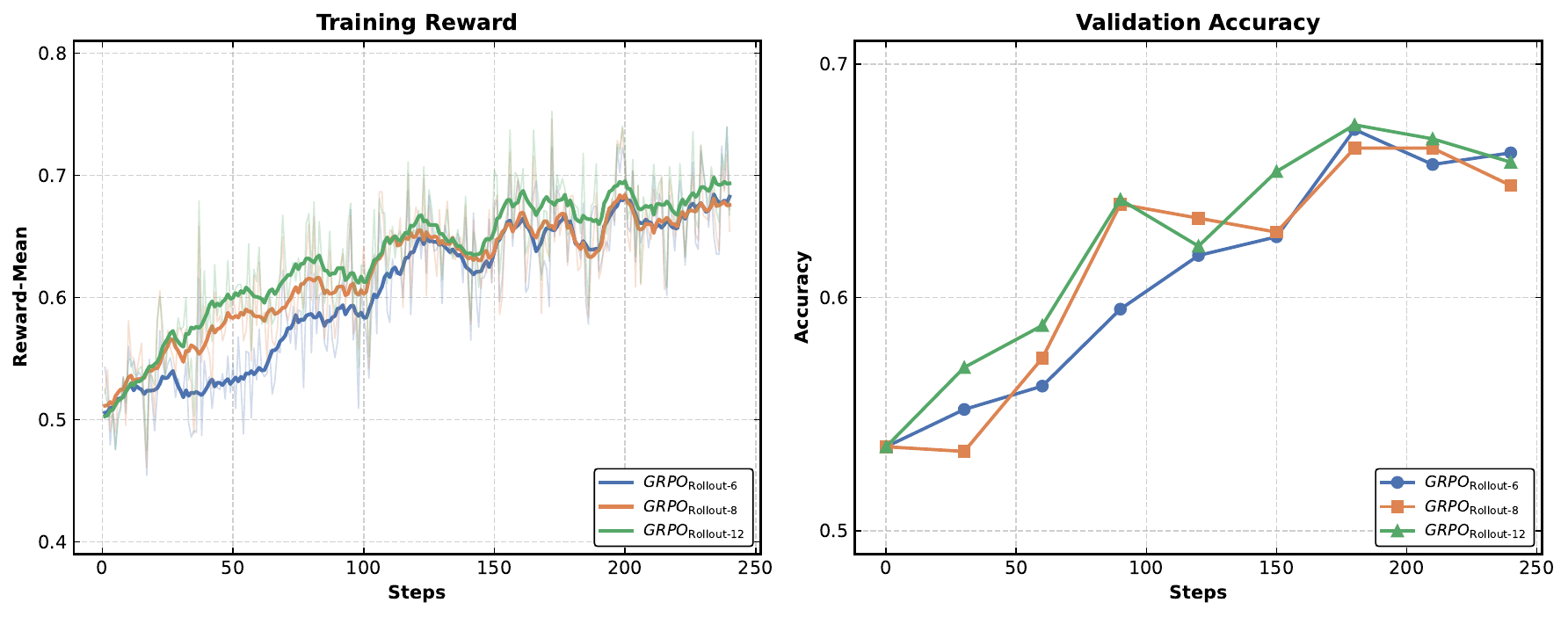}
  \caption{Training Dynamics of GRPO on DeepScaler-5K dataset}
  \label{fig:training_dynamics_deepscaler}
\end{figure}

In this section, we present the training dynamics of the GRPO baseline with different rollout budgets in Figures~\ref{fig:training_dynamics_deepmath} and~\ref{fig:training_dynamics_deepscaler}.

\section{Prompt Optimization Alleviates Advantage Collapse}
\label{app:advantage_collapse}

We measure advantage collapse at the level of individual rollout groups using the Self-Ref w/o-CRM variant on DeepMath-5K, with the policy frozen at the end of epoch 0 together with the $1{,}980$ hard samples mined during that epoch.
For each sample we draw $K=6$ rollouts under the bare query $x$ and another $K=6$ rollouts under the template-augmented query $\tilde x=\mathcal{T}(x,z)$, where the template $z$ is produced by GEPA at the end of epoch 0, and we average the per-sample group reward mean and variance over three buckets: all hard samples, all-incorrect samples ($\sum_{k=1}^{K} r_{ik}=0$, i.e., fully collapsed groups), and near-all-incorrect samples with $\sum_{k=1}^{K} r_{ik}/K \le 1/6$ (i.e., groups that retain at most one successful rollout).

\begin{table}[H]
\caption{\textbf{Advantage collapse on the hard samples of DeepMath-5K at the end of epoch 0.} Self-Ref w/o-CRM variant with $K=6$ rollouts per query. A sample is counted as \emph{Improved} if its average reward over the $K$ rollouts is higher when the template is applied than under the bare query. Variance and mean are averaged over the samples of each bucket; the two extreme buckets are nested inside the set of all hard samples.}
\label{tab:advantage_collapse}
\centering
\begin{small}
\begin{sc}
\setlength{\tabcolsep}{5pt}
\resizebox{\textwidth}{!}{%
\begin{tabular}{@{}lcccccc@{}}
\toprule
& & & \multicolumn{2}{c}{\textbf{Reward Variance}} & \multicolumn{2}{c}{\textbf{Reward Mean}} \\
\cmidrule(lr){4-5}\cmidrule(lr){6-7}
\textbf{Bucket} & \textbf{\# Samples (\%)} & \textbf{\# Improved (\%)} & w/o Template & w/ Template & w/o Template & w/ Template \\
\midrule
All hard samples & 1980 (100\%) & 740 (37.4\%) & 0.2494 & 0.2491 & 0.4765 & 0.5295 \\
All-incorrect ($\sum_{k=1}^{K} r_{ik}=0$) & 559 (28.2\%) & 202 (36.1\%) & 0 & 0.1205 & 0 & 0.1401 \\
Near-all-incorrect ($\sum_{k=1}^{K} r_{ik}/K \le 1/6$) & 745 (37.6\%) & 321 (43.1\%) & 0.0399 & 0.1617 & 0.0416 & 0.2029 \\
\bottomrule
\end{tabular}%
}
\end{sc}
\end{small}
\end{table}

Table~\ref{tab:advantage_collapse} shows that templates alleviate advantage collapse exactly where it is most severe.
The average reward rises in all three buckets, and more than one third of the samples in each bucket obtain a higher reward ($37.4\%$, $36.1\%$, and $43.1\%$).
In the fully collapsed bucket, the group reward variance is exactly $0$ without the template but reaches $0.1205$ with it, as $202$ groups escape complete failure and regain non-degenerate advantages; in the near-collapsed bucket, the variance rises from $0.0399$ to $0.1617$ ($4.1\times$) and the mean reward nearly quintuples ($0.0416\to0.2029$).
Notably, the aggregate variance over all hard samples is almost unchanged ($0.2494$ vs.\ $0.2491$), so the templates do not merely inflate the reward scale; they restore learning signals on the samples where advantage collapse would otherwise silence the policy.

\section{GEPA Iteration Dynamics}

\begin{figure}[h]
  \centering
  \includegraphics[width=1\linewidth]{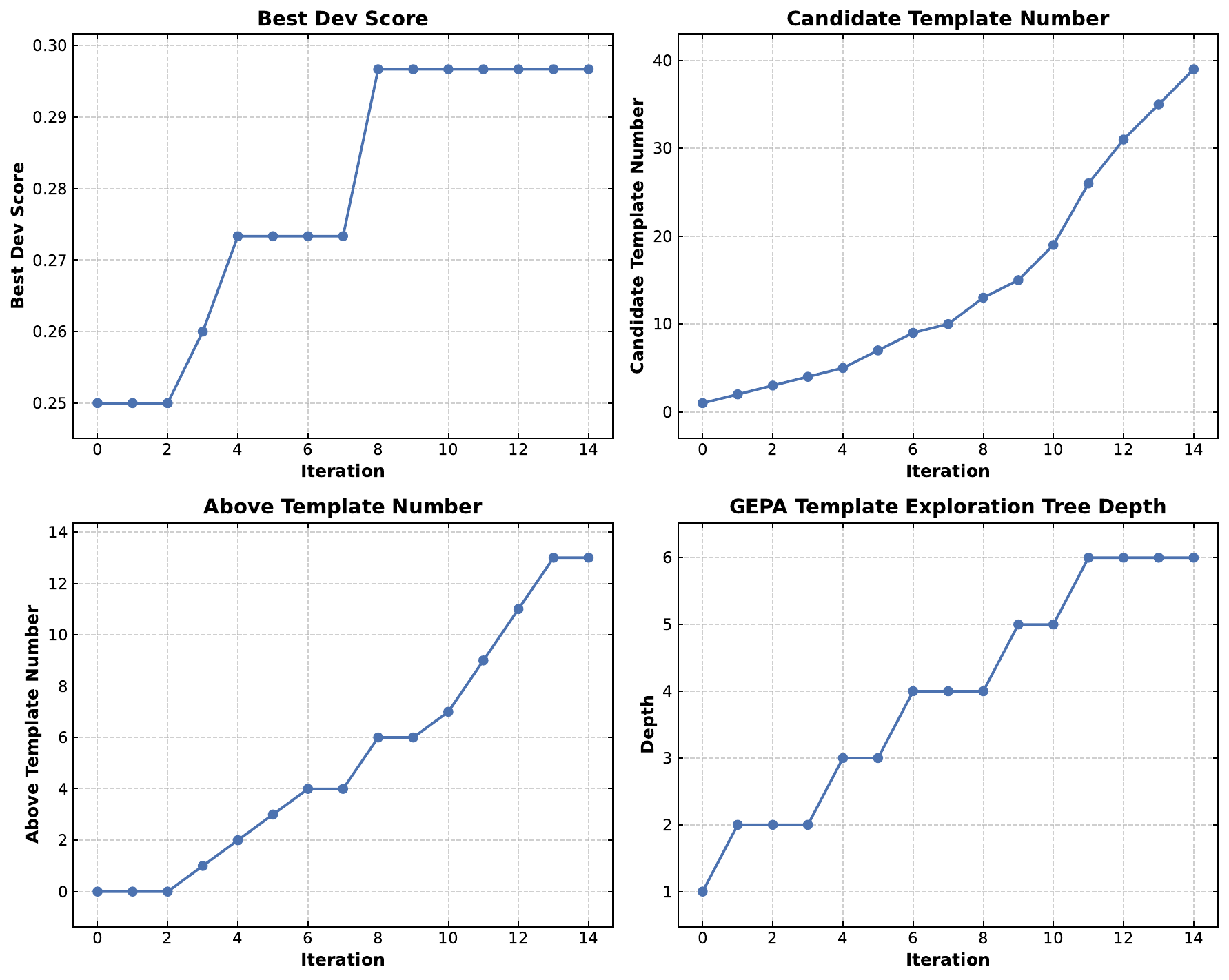}
  \caption{\textbf{Iteration Curve of GEPA}: Results of the Self-Ref variant on the DeepMath-10K dataset at epoch 1 during the GEPA iteration process. \textbf{Top left:} the change of the best template score. \textbf{Top right:} the number of candidate templates. \textbf{Bottom left:} the number of templates that outperform the empty template. \textbf{Bottom right:} the depth of the GEPA exploration tree.}
  \label{fig:gepa_iteration}
\end{figure}

Figure~\ref{fig:gepa_iteration} illustrates the internal optimization dynamics of GEPA, where the best template score exhibits a staircase-like ascending pattern: after two brief growth phases interspersed with plateaus, the score reaches its peak around iteration 8 and remains stable thereafter, indicating that GEPA reliably converges to a near-optimal prompt template within a limited number of iterations.

% \begin{figure}[t]
%   \centering
%   \includegraphics[width=1\linewidth]{figures/pass_k_and_hard_samples.pdf}
%   \caption{\textbf{Prompt Optimization Effectiveness.} \textit{Left:} Optimized prompts (triangles) consistently yield gains over the standard policy (circles) on both Pass@1 and Pass@6 metrics, demonstrating that GEPA effectively assists the model in bridging the performance gap. \textit{Right:} With this assistance, the model continuously conquers hard samples, resulting in a steady decline in the number of intractable instances throughout the training epochs. Data points are derived from the Teacher-Ref variant on the DeepScaler-5K dataset.}
%   \label{fig:pass_k_and_hard_samples}
% \end{figure}

\section{Why Stronger Reflection Does Not Guarantee a Better Mutation Operator}
\label{app:template_quality}

Section~\ref{subsec:main-results} shows that P$^2$O$_{\textnormal{Teacher-Ref}}$ trails P$^2$O$_{\textnormal{Self-Ref}}$ even though it reflects with a much stronger model, Kimi-K2. To isolate what the reflection model actually changes, we inspect the template set $\mathcal{Z}^{(1)}$ produced by the \emph{first} GEPA call, i.e., at the end of epoch 0, on DeepMath-5K with Qwen3-4B in the w/o-CRM configuration. At this point the two variants share identical policy weights and hard-sample sets $\mathcal{D}_{\text{hard}}^{(1)}$, and differ only in the template set returned by GEPA, so any gap in their dev-hard scores is attributable to the reflection model alone.

\begin{table}[H]
\caption{\textbf{Template quality versus downstream performance} DeepMath-5K, Qwen3-4B, w/o-CRM. Dev-hard scores are accuracies (\%) of the template set produced by GEPA at the end of epoch 0, averaged over all candidate templates (\emph{mean}) and taken at the best template (\emph{best}), where both variants evaluate on the same dev-hard split with the same policy weights. The last row reports the final six-benchmark average of the fully trained models (Table~\ref{tab:CRM_ablation_full}). $\Delta=\textnormal{Teacher-Ref}-\textnormal{Self-Ref}$.}
\label{tab:template_quality}
\centering
\begin{small}
\begin{sc}
\setlength{\tabcolsep}{6pt}
\begin{tabular}{@{}lccc@{}}
\toprule
Metric & Self-Ref & Teacher-Ref & $\Delta$ (Teacher $-$ Self) \\
\midrule
Mean dev-hard score & 16.2 & \textbf{24.4} & \gain{8.2} \\
Best dev-hard score & 26.7 & \textbf{31.7} & \gain{5.0} \\
Final downstream 6-benchmark AVG & \textbf{64.2} & 57.9 & \drop{6.3} \\
\bottomrule
\end{tabular}
\end{sc}
\end{small}
\end{table}

Table~\ref{tab:template_quality} exposes a clear inversion. On the dev split that GEPA itself optimizes, the teacher evolves the stronger template set by a wide margin: its mean dev-hard score is 24.4\% versus 16.2\%, and its best template reaches 31.7\% versus 26.7\%. The stronger reflection model thus does produce better mutations under the selection criterion, and it does so with the same policy and the same hard samples. Yet the model trained with this template set ends up $6.3$ points \emph{lower} (57.9\% vs.\ 64.2\%).

This inversion supports our hypothesis: what governs the benefit of prompt evolution is not the absolute strength of the reflection model or the dev score of the templates it produces, but whether the elicited behavior can be absorbed by the policy. Teacher-Ref templates earn their higher dev-hard scores with reasoning that the policy cannot yet produce on its own, so the trajectories they elicit lie farther outside the policy's current distribution, and context distillation transfers only a fraction of the prompt-induced gain into $\pi_\theta(\cdot\mid x)$. Self-Ref templates score lower on the dev split precisely because they remain close to the policy's own reasoning patterns: the successes they elicit are within reach, and each epoch of distillation converts a larger share of them into intrinsic capability.

\section{Why GEPA Finds Valid Strategies Rather Than Shortcuts}
\label{app:no_hacking}

One might suspect that GEPA improves the policy by leaking answer information into the evolved templates rather than by discovering genuine reasoning strategies. We rule out this shortcut hypothesis on four grounds.

\begin{enumerate}
\item \textbf{The exposed supervision is minimal.} Consider the Self-Ref w/o-CRM checkpoint on DeepMath-5K (Appendix~\ref{app:advantage_collapse}), whose $1{,}980$ hard samples are split into $300$ dev and $1{,}680$ train examples. GEPA reflects only on mini-batches drawn from the train split, and under the $10$K-rollout budget in practice fewer than $300$ distinct samples typically enter the reflection feedback at all---less than $15\%$ of the hard set. The remaining $85\%$ never expose their answers in any form, so even a verbatim encoding of those few hundred answers cannot explain the improvement on the untouched samples.

\item \textbf{The dev split filters out memorized content at selection time.} A mutation is kept only if it also improves the held-out $300$ dev hard samples, after which candidates are pruned by Pareto domination (Algorithms~\ref{alg:gepa_sub} and~\ref{alg:select_pareto}). A template that merely encodes the answer to a train sample yields no dev-side benefit and is discarded, so whatever enters the final template set must help on unseen samples---by construction, a transferable strategy.

\item \textbf{The template set is small and sample-agnostic.} The final covered set $\mathcal{Z}_{\text{covered}}$ (Algorithm~\ref{alg:greedy_assign}) typically contains only $15$--$25$ templates of at most $2{,}048$ tokens each, yet it serves thousands of hard samples per epoch. Assignment samples templates by dev coverage and average score and never retrieves or matches on question content, so an individual template has no structural channel for carrying per-question answers.

\item \textbf{The downstream evidence is inconsistent with a shortcut.} Under the memorization hypothesis the gains would be confined to the samples whose answers leaked into the templates, yet P$^2$O improves over GRPO by $21.5$ points on AIME24 and $11.0$ points on AIME25, far more than memorizing fewer than $300$ answers could plausibly produce. All six evaluation benchmarks are OOD---no template is ever optimized or scored on them---and after context distillation inference uses no template at all. Under these conditions, the gains can only be attributed to reasoning ability internalized in the model parameters.
\end{enumerate}

\section{Hard Sample Success-rate Threshold $\tau$ Ablation}
Table~\ref{tab:tau_ablation} presents the ablation results for the success-rate threshold $\tau$ defined in Eq.~\ref{eq:hard_rate}. With $K=6$, the tested rates $\{1/6,2/6,4/6\}$ correspond to hard-sample cutoffs of fewer than $\{1,2,4\}$ successful rollouts. Compared with $\tau=1/6$, which retains only fully incorrect samples, our default $\tau=2/6$ improves average accuracy by 2.1 points. The lower threshold yields too few samples for GEPA to iterate over, causing prompt optimization to converge to a suboptimal local optimum. Increasing the rate to $\tau=4/6$ performs comparably to the default (64.1 vs.\ 64.2 average), suggesting robustness to moderate threshold changes.

\begin{table}[H]
\caption{\textbf{Hard Sample Success-rate Threshold $\tau$ Ablation.} Results on DeepMath-5K using the Self-Ref w/o-CRM variant with $K=6$. Thresholds are reported as rates. Our default $\tau=2/6$ treats samples with at most one correct rollout as hard.}
\label{tab:tau_ablation}
\centering 
\begin{small}
\begin{sc}
\setlength{\tabcolsep}{4pt}
\resizebox{\textwidth}{!}{
\begin{tabular}{lccccccc}
\toprule
Model & AIME24 & AIME25 & AMC & MATH500 & Minerva & Olympiad & AVG \\
\midrule
Qwen3-4B-P$^2$O$_{\tau=2/6}$ & 56.7 & \textbf{45.4} & 86.9 & \textbf{92.4} & 40.4 & \textbf{63.1} & \textbf{64.2} \\
Qwen3-4B-P$^2$O$_{\tau=1/6}$ & 50.0 & 42.5 & 87.3 & 90.2 & 40.8 & 61.6 & 62.1 \\
Qwen3-4B-P$^2$O$_{\tau=4/6}$ & \textbf{57.3} & 44.6 & \textbf{88.0} & 91.6 & \textbf{42.6} & 60.4 & 64.1 \\
\bottomrule
\end{tabular}%
}
\end{sc}
\end{small}
\end{table}

\section{Context Ratio Mask Ablation}
\label{app:CRM_ablation}

Table~\ref{tab:CRM_ablation_full} reports the complete benchmark-wise results of the ablation summarized in Section~\ref{subsec:ablations}. For every training set, backbone, and reflection source, we compare P$^2$O against its \textit{w/o-CRM} counterpart, where the Context Ratio Mask of Eq.~\ref{eq:p2o_objective} is removed while hard-sample mining, template assignment, and context distillation are kept identical.

\begin{table}[H]
\caption{\textbf{Full Results of the Context Ratio Mask Ablation.} Benchmark-wise accuracy (\%) of P$^2$O and its w/o-CRM counterpart with the Qwen3-4B and Qwen3-8B backbones, on the DeepMath-5K and DeepScaler-5K training sets and the two reflection sources. AVG averages the six benchmarks.}
\label{tab:CRM_ablation_full}
\centering
\begin{small}
\begin{sc}
\setlength{\tabcolsep}{4pt}
\resizebox{\textwidth}{!}{%
\begin{tabular}{lccccccc}
\toprule
Model & AIME24 & AIME25 & AMC & MATH500 & Minerva & Olympiad & AVG \\
\midrule
\rowcolor{gray!15} \multicolumn{8}{c}{\textbf{DeepMath-5K}} \\\midrule
Qwen3-4B-P$^2$O$_{\textnormal{Self-Ref}}$ & 55.3 & 40.0 & 86.2 & 91.2 & 43.0 & 62.4 & 63.0 \\
Qwen3-4B-P$^2$O$_{\textnormal{Self-Ref, w/o-CRM}}$ & 56.7 & 45.4 & 86.9 & 92.4 & 40.4 & 63.1 & 64.2 \\
\midrule
Qwen3-4B-P$^2$O$_{\textnormal{Teacher-Ref}}$ & 46.5 & 32.5 & 82.3 & 88.6 & 41.5 & 59.3 & 58.5 \\
Qwen3-4B-P$^2$O$_{\textnormal{Teacher-Ref, w/o-CRM}}$ & 44.6 & 34.4 & 81.9 & 90.2 & 36.0 & 60.0 & 57.9 \\
\midrule
Qwen3-8B-P$^2$O$_{\textnormal{Self-Ref}}$ & 57.1 & 41.7 & 86.7 & 92.2 & 43.0 & 63.4 & 64.0 \\
Qwen3-8B-P$^2$O$_{\textnormal{Self-Ref, w/o-CRM}}$ & 51.0 & 39.2 & 86.9 & 89.6 & 40.8 & 60.3 & 61.3 \\
\midrule
Qwen3-8B-P$^2$O$_{\textnormal{Teacher-Ref}}$ & 53.4 & 42.3 & 88.8 & 91.2 & 40.8 & 63.9 & 63.4 \\
Qwen3-8B-P$^2$O$_{\textnormal{Teacher-Ref, w/o-CRM}}$ & 49.2 & 40.0 & 85.5 & 90.8 & 39.7 & 60.9 & 61.0 \\
\midrule
\rowcolor{gray!15} \multicolumn{8}{c}{\textbf{DeepScaler-5K}} \\\midrule
Qwen3-4B-P$^2$O$_{\textnormal{Self-Ref}}$ & 59.0 & 42.2 & 88.4 & 92.0 & 42.6 & 62.1 & 64.7 \\
Qwen3-4B-P$^2$O$_{\textnormal{Self-Ref, w/o-CRM}}$ & 51.5 & 42.1 & 88.1 & 91.0 & 40.1 & 61.8 & 62.4 \\
\midrule
Qwen3-4B-P$^2$O$_{\textnormal{Teacher-Ref}}$ & 57.7 & 40.2 & 87.0 & 91.6 & 43.4 & 64.0 & 64.0 \\
Qwen3-4B-P$^2$O$_{\textnormal{Teacher-Ref, w/o-CRM}}$ & 59.8 & 49.4 & 92.2 & 91.4 & 36.4 & 62.2 & 65.2 \\
\midrule
Qwen3-8B-P$^2$O$_{\textnormal{Self-Ref}}$ & 61.7 & 45.4 & 89.5 & 91.6 & 41.5 & 63.6 & 65.6 \\
Qwen3-8B-P$^2$O$_{\textnormal{Self-Ref, w/o-CRM}}$ & 53.1 & 41.3 & 86.1 & 91.4 & 42.3 & 59.0 & 62.2 \\
\midrule
Qwen3-8B-P$^2$O$_{\textnormal{Teacher-Ref}}$ & 55.5 & 45.0 & 85.7 & 91.2 & 43.0 & 59.6 & 63.3 \\
Qwen3-8B-P$^2$O$_{\textnormal{Teacher-Ref, w/o-CRM}}$ & 50.0 & 37.9 & 84.5 & 90.6 & 42.3 & 58.2 & 60.6 \\
\bottomrule
\end{tabular}%
}
\end{sc}
\end{small}
\end{table}

\section{Case Study}

\begin{figure}[h]
  \centering
  \includegraphics[width=1\linewidth]{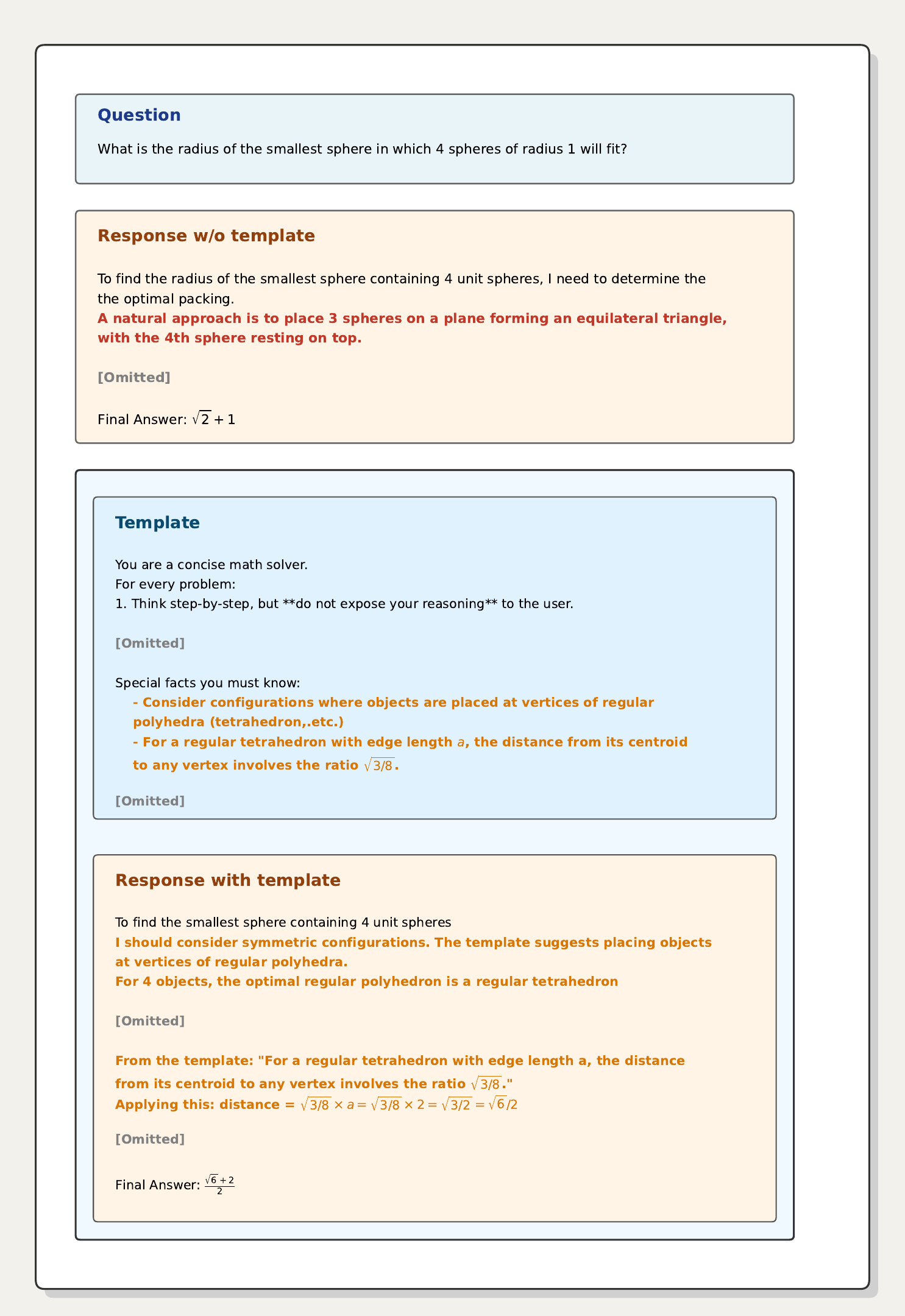}
  \caption{Qualitative Analysis: Overcoming Local Optima in Geometric Reasoning.}
  \label{fig:case_study}
\end{figure}

To further demonstrate how P$^2$O overcomes the exploration bottleneck, we analyze a representative ``hard sample'' from the training process in Figure~\ref{fig:case_study}. The problem asks for the radius of the smallest sphere enclosing four unit spheres.

\paragraph{Failure of the Base Policy.} 
Without guidance, the model falls into a common cognitive trap: it defaults to a visually intuitive but mathematically suboptimal configuration. As shown in the generated trace, the model attempts to place ``3 spheres on a plane... with the 4th sphere resting on top.'' This corresponds to a localized packing that yields a parent sphere radius of $1+\sqrt{2}$, failing to minimize the volume.

\paragraph{Mechanism of the Optimized Prompt.}
The template evolved by GEPA acts as a targeted intervention in the model's latent search space. It does not merely encourage the model to ``think harder''; rather, it injects specific domain knowledge---specifically the concept of the regular tetrahedron and the precise mathematical constant for its centroid-to-vertex distance ($\sqrt{3/8}$).
This prompt effectively prunes the invalid search space (planar configurations) and steers the reasoning trajectory toward the global optimum (tetrahedral packing), allowing the model to derive the correct radius of $1 + \frac{\sqrt{6}}{2}$.

\end{document}